\documentclass[runningheads]{llncs}
\usepackage[utf8]{inputenc}
\usepackage{booktabs}
\usepackage{graphicx}
\usepackage{hyperref}
\usepackage{authblk}
\usepackage{amsmath}
\usepackage{subcaption}
\usepackage{todonotes}
\usepackage[misc,geometry]{ifsym}

\title{Generating Human-Like Movement: \\ A Comparison Between Two Approaches Based on Environmental Features}

\author{A. Zonta \Letter\inst{1, 2} \and
S. K. Smit\inst{2} \and
A.E. Eiben\inst{1}}

\institute{Department of Computer Science \\ Vrije Universiteit Amsterdam, De Boelelaan 1111, 1081 HV, \\ Amsterdam, The Netherlands \\ \email{{a.zonta, a.e.eiben}@vu.nl} \and
Modelling, Simulation \& Gaming \\ TNO, The Hague, The Netherlands \\
\email{{alessandro.zonta, selmar.smit}@tno.nl}}

\authorrunning{A. Zonta et al.}
\titlerunning{Generating Human-Like Movement: A Comparison}

\begin{document}

\maketitle

\begin{abstract}
Modelling realistic human behaviours in simulation is an ongoing challenge that resides between several fields like social sciences, philosophy, and artificial intelligence. 
Human movement is a special type of behaviour driven by intent (e.g. to get groceries) and the surrounding environment (e.g. curiosity to see new interesting places). 
Services available online and offline do not normally consider the environment when planning a path, which is decisive especially on a leisure trip.
Two novel algorithms have been presented to generate human-like trajectories based on environmental features.
The Attraction-Based A* algorithm includes in its computation information from the environmental features meanwhile, the Feature-Based A* algorithm also injects information from the real trajectories in its computation.
The human-likeness aspect has been tested by a human expert judging the final generated trajectories as realistic.
This paper presents a comparison between the two approaches in some key metrics like efficiency, efficacy, and hyper-parameters sensitivity. 
We show how, despite generating trajectories that are closer to the real one according to our predefined metrics, the Feature-Based A* algorithm fall short in time efficiency compared to the Attraction-Based A* algorithm, hindering the usability of the model in the real world.
\end{abstract}

\keywords{Trajectory Generation  \and Human-likeness \and Human Trajectories \and Generative Methods}

\section{Introduction}
\label{section:intro}
% simulation - 
The importance of simulations is increasing in the Artificial Intelligence world.
From video games trying to realise open-world environments as realistic as possible to virtual gyms for testing robot and flight simulators to training pilots, simulations are an easy and cheap way to test the skills of people and robots without causing harm to themselves and the environment.
Focusing on urban simulations, the ability to model the trajectories of people and make them as realistic as possible would improve the gain that people can have on training with these simulations or lower the effort needed to create a realistic world and hence have a better experience.
Considering a training task in a virtual environment for a law enforcement officer, normal and hostile behaviours do coexist together.
The task to spot different behaviours that might indicate hostile conditions is made more realistic by the introduction of authentic pedestrians moving around the map instead of scripted ones.
From optimisation of traffic flows, the management of crowds around public spaces or after big events, securing the population from possible terrorist threats, or even military training; a model that represents how a human behaves under certain conditions is a subject attracting a lot of researchers.

% environmental features
When humans move, especially for leisure, they are driven by their intent (e.g. to get groceries) and the surrounding environment (e.g. curiosity to see new interesting places). 
During holiday planning, several online resources are consulted to find the most fascinating points in the area, and a plan is made to maximise the number of attractive locations visited.
This process is feasible with several online services when we consider cities, monuments and seaside resorts as environmental features.
Unfortunately, to the best of our knowledge, there are no services that let you give a preference on the type of area someone wants to visit, i.e. higher importance to forest area compared to an industrial area.
Being able to have a service that accepts this kind of preference system would provide a new opportunity for exploring areas around the world and, at the same time, would also improve the variety of behaviours expressible in urban simulations.
Having a diverse set of behaviours in simulation is a decisive ingredient towards making them realistic since their homogeneity makes them easy to catch and, at the same time, do not represent what happens in the real world.

% the need to be human like - and automatic
Designing a model that can perfectly replicate the real paths is not required and probably even not desirable, since all the paths close to the real ones could be considered realistic, due to all the possible natural interference caused by moving in the real world, i.e. avoiding obstacles, traffic lights, and crowded places.
The objective of having real paths, real in a way a human could have generated it, and not a perfect match with the recorded one makes the task more complex, and especially finding a metric to expresses automatically the human-likeness is very difficult.

A way to solve the problem and assure the human-likeness of these trajectories is employing a human expert to check and verify them.
\cite{Zonta2019} and \cite{Zonta2020} manages to achieve so with two modifications of the well known A* algorithm.
Given the similarity between the path planning problem and generation of trajectories task, the famous path planner algorithm A* was chosen to become the root of two novel ways to generate trajectories.
A way to describe the environment was presented in \cite{Zonta2017}, which suits perfectly the needs of the A*  algorithm given its matrix/graph representation.
Environmental features are easily downloadable from online map services and codified into a map, which is inspired by the concept of electrical charges.
An individual moving along the map is seen as an electrically-charged particle, with features having an opposite type of electrical charge to attract the individual.
The map also includes the routing system of the area under study, forcing the individual to move along it, making the movements realistic.
Information about the environment is injected into the A*  algorithm thanks to this representation, leading to a more realistic generation of trajectories. 

\cite{Zonta2020} starts a short comparison between the two algorithms, but more in-depth analysis is needed to understand the advantages of using one version over the other.
In this paper, we aim to conduct such analysis following several key points summarised in the following statement: 
``What are the differences between the two algorithms regarding:"
\begin{itemize}
    \item Efficiency 
    \item Efficacy 
    \item Preference between total distance or attractiveness of the environment
    \item Sensitivity to environmental features
\end{itemize}
Efficiency is a measurable concept, quantitatively determined by the ratio of useful work to resources (i.e. processor and storage) expended. 
In other words, the ratio of the output to the input of a given system.
Specifically, we compare time efficiency and sampling efficiency.
Time efficiency is defined as the amount of time that an algorithm needs to reach a certain level of performance. 
Sample efficiency denotes the amount of experience that an algorithm needs to generate in an environment (e.g. the number of actions it takes ) during training to reach a certain level of performance.
Efficacy is defined as the ability to perform a task to a satisfactory or expected degree.
In our case, it defines how good are the trajectories based on predefined metrics.
One of the hyper-parameters needed by the two algorithms is the degree of preference towards the attractiveness of an area and willingness to focus towards the distance remaining to generate.
This can change drastically how the final trajectories will look like, considering the example of a trip where you want to visit one or two interesting locations and, at the same time, travel as far as possible.
Hyper-parameters are also the set of weights related to the environmental features able to tweak the trajectory preferences.
Keeping the example of the trip, this would allow visiting more wooded area than industrial ones.
A more in-depth analysis of the influence of these hyper-parameters is presented in this paper.\\

This paper is structured as follows: In Section \ref{section:related} we explain some related work, meanwhile, in Section \ref{section:world} we explain the world representation we used to map the interesting points in the area with the routing system.
Section \ref{section:features_representation} describes the features used by one of the versions of the A*  algorithm and how we derived them, with Section \ref{section:models} describing the two A* algorithm.
After the the dataset containing the real trajectories explained in Section \ref{section:dataset}, and the experimental setup explained in Section \ref{section:experimental_setup}, the results are presented in section \ref{section:results} together with an in depth discussion.
Our concluding remarks are summarised in section \ref{section:conclusion}.

\section{Related Work}
\label{section:related}
Given the two main areas included in the paper, the same division is followed when presenting the related work, starting from modelling trajectories to the various modification to the A* algorithm existing in the literature.

\subsection{Modelling Trajectories}
    Several methods have been studied to model and predict possible destinations or entire trajectories.
    \cite{Zhou2012} 
    uses a Mixture model of Dynamic pedestrian-Agents to learn the collective behaviour and patterns of pedestrians in crowded scenes.
    With this method, they show good performance in the prediction of the next movement of a person in a crowd.
    \cite{Luber2010} models human behaviour using hand-crafted functions such as the idea of Social Forces.
    Social forces define all the forces that influence every person during their movement, e.g. motivation towards the goal, obstacles and other people along their path.
    With the help of a Kalman filter-based multi-hypothesis tracker, they show the importance of these forces in modelling the behaviours of a crowd of people.
    \cite{Zheng2014} proposed a novel concept, called gathering, which is a trajectory pattern modelling various group incidents such as celebrations, parades, protests, traffic jams, etc. 
    Batch and real-time analysis of trajectories dataset are presented to show the power of the concept developed. 
    A novel representation method of trajectory data, called Activity Description Vector (ADV), is presented by \cite{JorgeAzorin-LopezMarceloSaval-CalvoAndresFuster-Guillo2013} based on the number of times a person is in a specific area in the scenario under analysis and the movement that the person is performing.
    This representation has been used as input for different classifiers in order to recognise peoples behaviour in a shopping centre.
    The Kalman filter model is also applied by \cite{Ellis2009} to predict changes in target position based on their current position and with the help of a previously identified cluster of similar trajectories.
    
    In contrast to the last papers, several publications moved towards the direction of a model able to learn, based on data, the idea of forces by itself without pre-defining them.
    Reference \cite{Alahi2016} uses an LSTM to predict the future positions of pedestrians and adjust accordingly its path to avoid collisions, \cite{Zegers2003} uses a novel neural network-based solution procedure for the trajectory generation able to emulate any desired trajectory behaviour irrespective of its complexity.
    \cite{B2017} employs recurrent neural networks (RNNs) in order to predict a user’s destinations from their partial trajectories estimating the transition probabilities for the next time step.
    \cite{Zonta2017} presents a method to model the intended destination of a subject in real-time, based on a trace of position information and prior knowledge of possible destinations.
    The method models the certainty of each Point of Interest by means of a virtual charge, resulting in an artificial potential field that reflects the current estimate of the subject’s intentions. 
    The virtual charges are updated as new information about the subject’s position arrives.
    \cite{Zonta2019} shows how different machine learning algorithms trained on real data perform poorly compared to a variation of the A* path planning algorithm in the task of generating human-like trajectories based on environmental features. 
    
    The problem of generation feasible trajectories has been tackled also in the robotic domain, with \cite{Howard2009} presenting a highly generic approach to trajectory generation for mobile robots of different mobility and shape characteristic.
    In order to generate a smooth, continuous path on flat terrain, an arbitrary number of constraints involving position, heading, linear and angular velocities, and curvature have to be satisfied.
    
\subsection{A* modifications}
    Several modifications have been proposed for the well-know A* algorithm.
    \cite{Duchon2014} presents modifications and improvements of the A* algorithm for path planning of a mobile robot based on a grid map.
    They introduce an every angle search for the next cell, a method to reduce the amount of examined cells and an approach to reduce the computational time of the algorithm.
    \cite{Fernandes2015} presents an extension of the A* algorithm in a cell decomposition, where besides its position, the orientation of the platform is also considered when searching for a path. 
    This is achieved by constructing different layers of orientations and only visiting neighbour layers when searching for the lowest cost. 
    This results in an algorithm capable of generating smooth, feasible paths for an any-shape mobile robot, which takes into account orientation restrictions, with the final aim of navigating close to obstacles. 
    \cite{Cheng2014} designed an improved A* algorithm to increase searching efficiency and precision in a large parking guidance system task.
    The improvement is accomplished by designing a hierarchical A* algorithm which divides the path searching into a few processes, the optimal solution for each process is found out, and then the global optimal solution is obtained. 
    \cite{Zonta2019} shows how a modified version of the well known A* algorithm outperforms different machine learning algorithms by computed evaluation metrics and human evaluation in the task of generating bike trips in the area around Ljubljana, Slovenia.
    The information from an artificial potential field describing the area under interest is included in the computation of the heuristic.
    \cite{Wang2018a} proposes an improved algorithm in consideration of the orientation constraint of a marine robot for A* algorithm.
    They state that paths generated via this proposed method are more appropriate than classic A*’s in practical application.

\section{World Representation}
\label{section:world}
In order to generate trajectories that match the routing system, i.e. the scheme of interconnecting lines and points that represent the streets or roads existing in the real world, a virtual navigable representation is required. 
In this paper, we chose the Artificial Potential Field (APF) \cite{Zonta2017} idea as a base to describe the routing system.
Points of Interest (POIs) need to be defined in order to codify all the environmental features of the area under analysis.

\subsection{POIs}
\begin{figure}[!t]
	\begin{subfigure}{.48\textwidth}
		\centering 
		\includegraphics[scale=0.4]{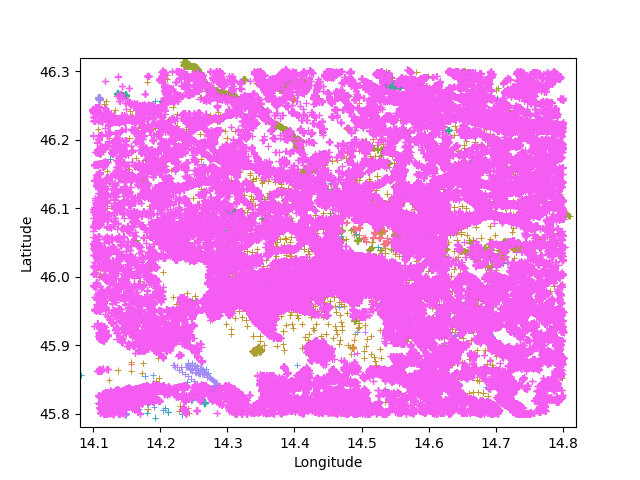} 
		\caption{Natural objects locations on map}
		\label{img:natural_locations}
	\end{subfigure}
	\begin{subfigure}{.48\textwidth}
		\centering 
		\includegraphics[scale=0.4]{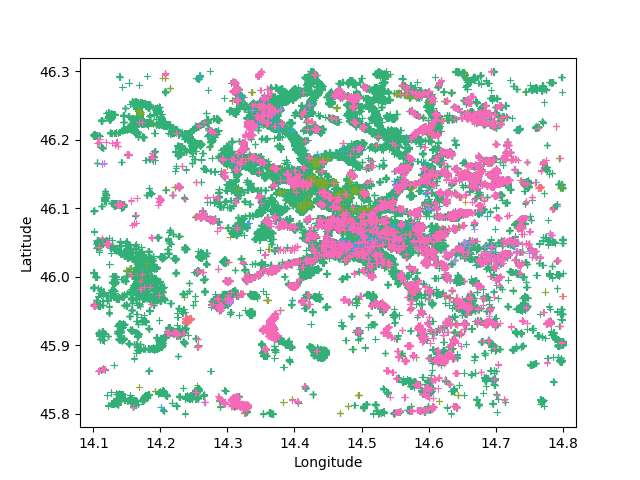} 
		\caption{Building objects locations on map}
		\label{img:building_locations}
	\end{subfigure}
	\vskip\baselineskip
	\begin{subfigure}{.48\textwidth}
		\centering 
		\includegraphics[scale=0.4]{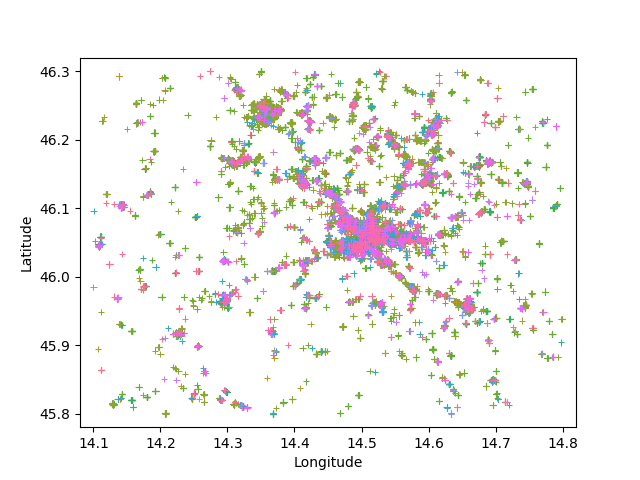} 
		\caption{Amenity objects locations on map}
		\label{img:amenity_locations}
	\end{subfigure}
	\begin{subfigure}{.48\textwidth}
		\centering 
		\includegraphics[scale=0.4]{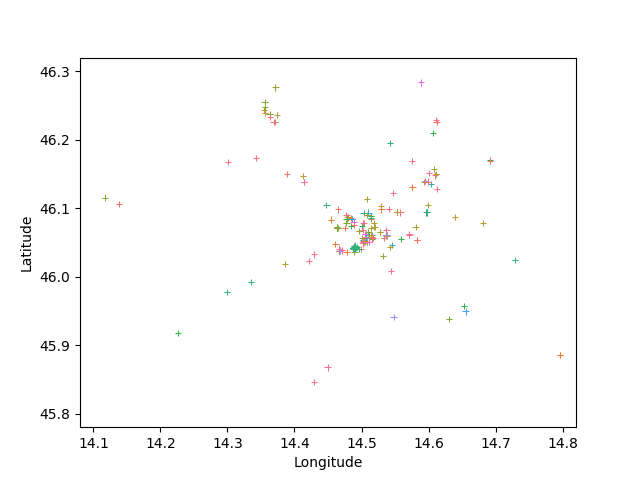} 
		\caption{Office objects locations on map}
		\label{img:office_locations}
	\end{subfigure}
	\vskip\baselineskip
	\begin{subfigure}{.48\textwidth}
		\centering 
		\includegraphics[scale=0.4]{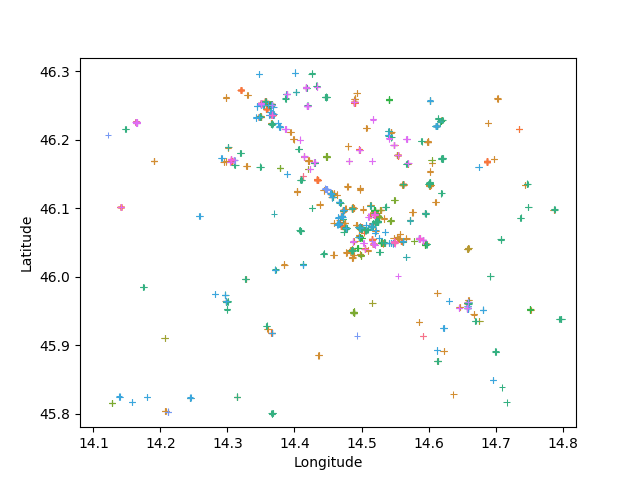} 
		\caption{Sport objects locations on map}
		\label{img:sport_locations}
	\end{subfigure}
	\begin{subfigure}{.48\textwidth}
		\centering 
		\includegraphics[scale=0.4]{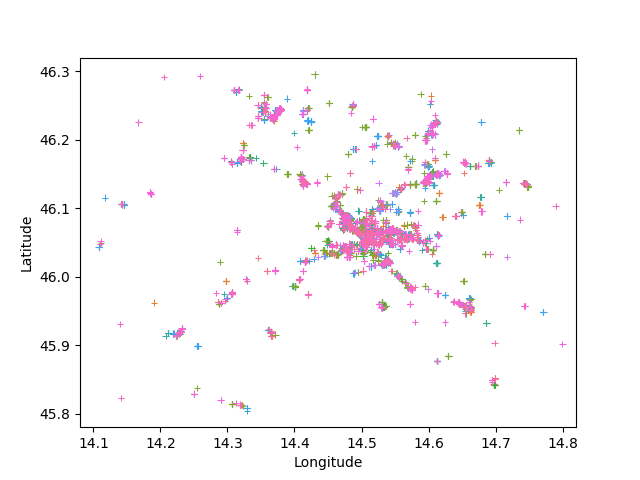} 
		\caption{Shop objects locations on maps}
		\label{img:shop_locations}
	\end{subfigure}
	\caption{Distribution of all the objects downloaded. Different colour correspond to different sub-typology of the tags.}
	\label{img:locations}
\end{figure}

    In order to generate human-like trajectories, the system defines POIs in the area that might be interesting for a real human.  
    As POIs we define all the different categories of constructions/objects that are situated in the area of interest.
    All the information are freely downloadable from Open Street Maps\footnote{\url{https://www.openstreetmap.org/}} (OSM).
    The six most descriptive classes (tags) are chosen: \textit{\emph{B}uilding}, \textit{\emph{A}menity}, \textit{\emph{N}atural}, \textit{\emph{O}ffice}, \textit{\emph{Sh}op}, and \textit{\emph{Sp}ort}.
    For every location labelled with one of these tags, we obtained their GPS coordinates.
    If the location is not represented by a single node, i.e. unique GPS position, but by a polygon, i.e. border of the building/park, the centroid of the polygon is taken as coordinates of the location. 
    Table \ref{tab:poi_details} shows some details about the objects downloaded and Fig. \ref{img:locations} shows the distribution of the objects in the area under interest.
    As clearly visible, the presence of natural objects is dominant compared to the other kind of objects, with sport, amenity, shop, and office mainly localised in what seems like a city centre, which shape is recognisable from the Fig. \ref{img:amenity_locations}, \ref{img:sport_locations}, and \ref{img:shop_locations}.
    Different colours on the figures represent sub typology of the tag, which we all collapsed on their respective top tags. 
    \begin{table}
        \centering 
    	\caption{Number of downloaded objects divided by tags}
    	\label{tab:poi_details}
    	\begin{tabular}{c|cccccc}
    		\toprule
    		& Sport & Shop &Office&Building&Natural& Amenity\\
    		\midrule
    		elements & 37220 & 4637& 821&496925&968541&29099\\
    		\bottomrule
    	\end{tabular}
    	\vspace*{-3mm}
    \end{table}
    
\subsection{APF}

    \begin{figure}[t]
    \begin{subfigure}{.48\textwidth}
        \includegraphics[scale=0.28]{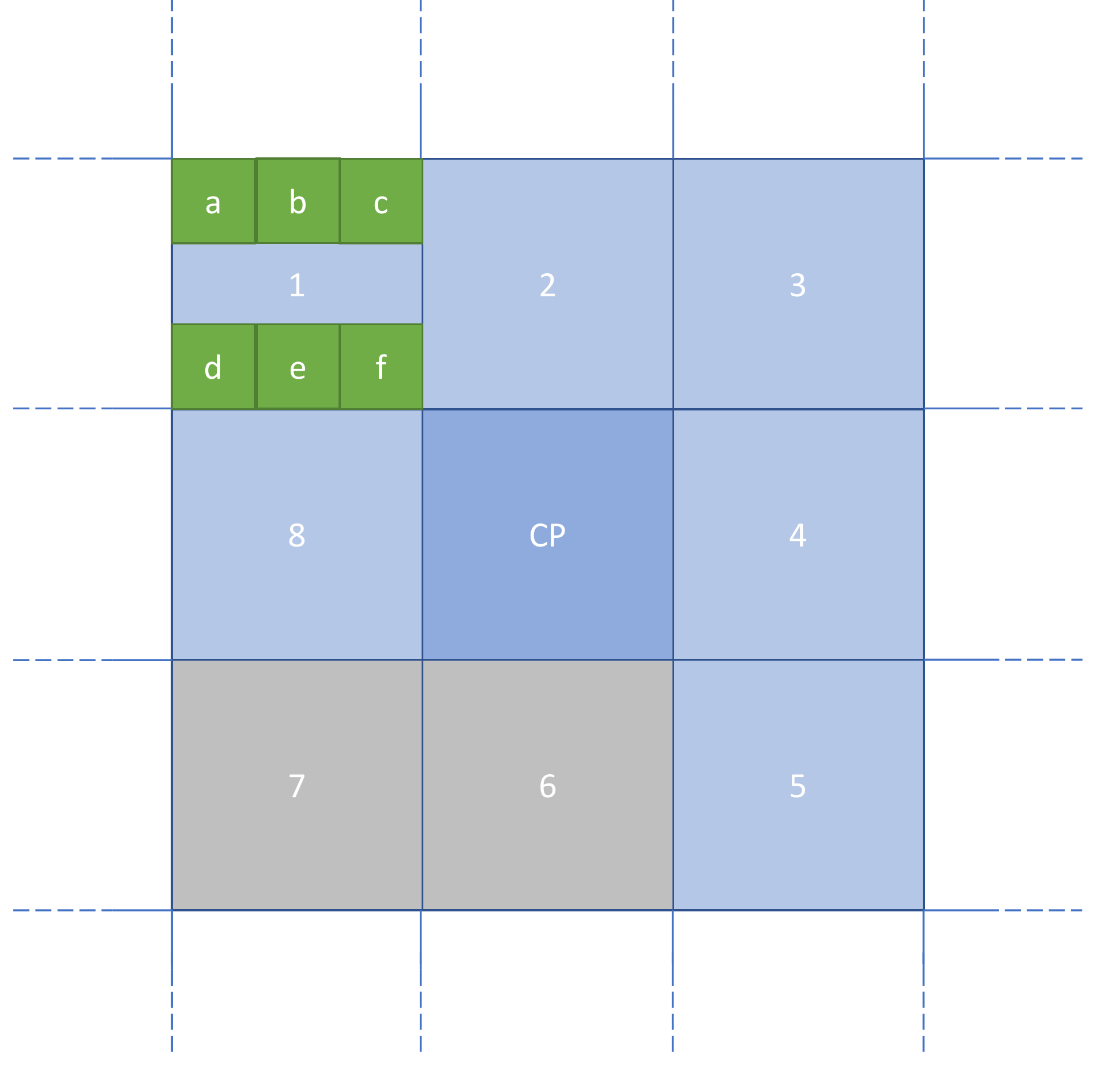} 
    	\caption{Matrix visualisation.}
    	\label{img:grid}
    \end{subfigure}
    \begin{subfigure}{.48\textwidth}
        \includegraphics[scale=0.45]{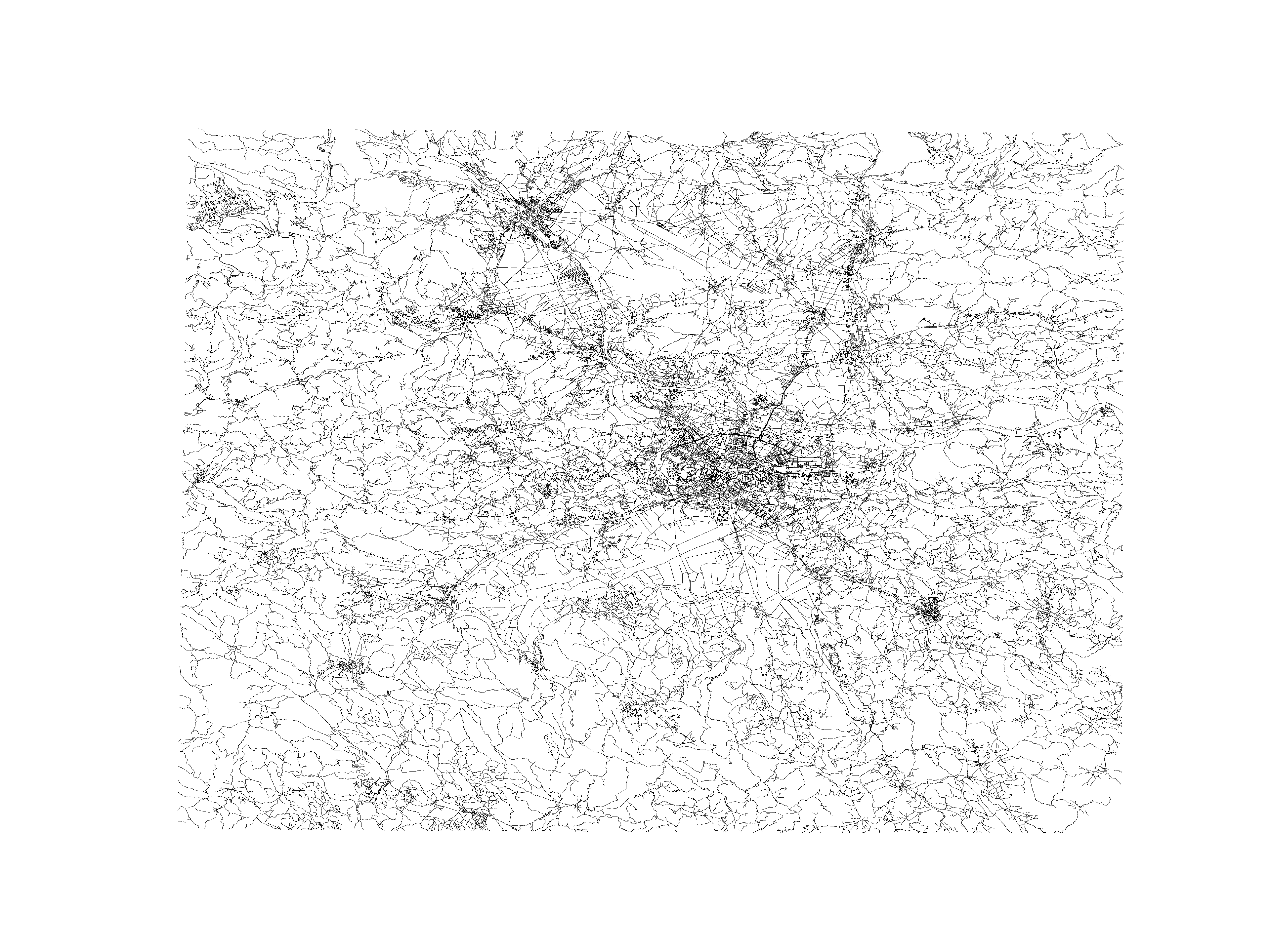} 
    	\caption{Heat map.}
    	\label{img:apf}
    \end{subfigure}
    \caption{Figure \ref{img:grid} shows the matrix visualisation of the current position (CP) and the neighbours cells. Every cell contains six values describing the charge of the six tags. The colour of the cell describes the presence of a road or not. Figure \ref{img:apf} shows the heat map of the APF created reading the road system map.}
    \end{figure}
    
    As a consequence of the dataset (see Section \ref{section:dataset}) containing data from Slovenia, we downloaded from OSM the area surrounding Ljubljana.
    The data is represented as a matrix, where only the cells defining a road have a defined value $V$ meanwhile, all the others have 0 to indicate empty spaces, as visible in Fig. \ref{img:apf}.
    Each one of the six tags carries a charge $q$ that indicates how attractive that family of objects is to the virtual agent, i.e. the system generating the trajectories.
    This structure defines an APF where there are several points, i.e. the POIs, attracting another point, i.e. the tracked person.
    This APF serves as a model of the agent preferences and intentions with the highest charged tag representing the most likely objects the users are attracted to.

    Equation \ref{eq:coulomb} is used by the model to compute the attraction of the objects from a defined cell of the matrix:
    \begin{equation}\label{eq:coulomb}
    Q = \sum_{k=0}^{tags}\sum_{i=0}^{O_k}\frac{\left | q_i \right |}{d^2} 
    \end{equation}
    The formula corresponds to Coulomb's law equation that defines the amount of force (attraction) \textit{Q} created by charges \textit{q} at a certain distance \textit{d}.
    Per every tag $k$ the attraction is computed considering all the POIs $O_k$.

    \figurename{~\ref{img:grid}} shows a representation of the  APF which includes the current position (CP) and the neighbours' cells. 
    The colour of the cell represents if a road is present (blue) or not (grey).
    For every cell, we have six values, representing the attraction that all the points from the different tags excerpt in that location, computed using Eq. \ref{eq:coulomb}.
    The order the system considers the neighbours' cells is clockwise and it is represented by the number on the cells.

\section{Features Representation}
\label{section:features_representation}
Designing a model that can perfectly replicate the real paths is not required and probably even not desirable, since all the paths close to the real ones could be considered realistic, due to all the possible natural interference caused by moving in the real world.
A starting point to generate trajectories with characteristics that resemble the real ones is to inject some information about the real trajectories into the generating methods.
At the same time, we cannot inject too much information otherwise the system loses generality and ends up generating the same trajectories present in the real dataset.
The idea behind the usage of these features derives by our interpretation of human-likeness.
To be considered realistic and human-like, the trajectories must follow particular parameters we define and being certified by a human expert.
The parameters we defined derived from the real trajectories, under the form of features.
Given the assumption the real trajectories can be all defined human-like, we can derive some features to boost our models in the task to generate new human-like trajectories.
We selected three features we found relevant to describe the trajectories: 
\begin{itemize}
    \item \textbf{total length}: how many time steps the trajectory is long. 
    \item \textbf{curliness}: a value that represents how straight or curly the trajectory is.
    \item \textbf{distance} of the \textbf{farthest point} from the \textbf{start}: how much the trajectories extend from the starting point.
\end{itemize}
Meanwhile, total length and distance of the farthest point are straightforward in their computation, curliness is a derived feature.
Considering a single trajectory, for every point in it, a binary vector of the movement to reach the next point is created, with 0 describing the neighbours' cell where the trajectory is not going and 1 describing the next position.
Once we obtained the movement vector for the entire trajectory, the Euclidean distance between them is computed and then averaged to obtain a single value per trajectory.

\begin{figure*}
 	\raggedright
 	\begin{subfigure}[b]{0.99\textwidth}
 	 	\centering
 		\includegraphics[scale=0.40]{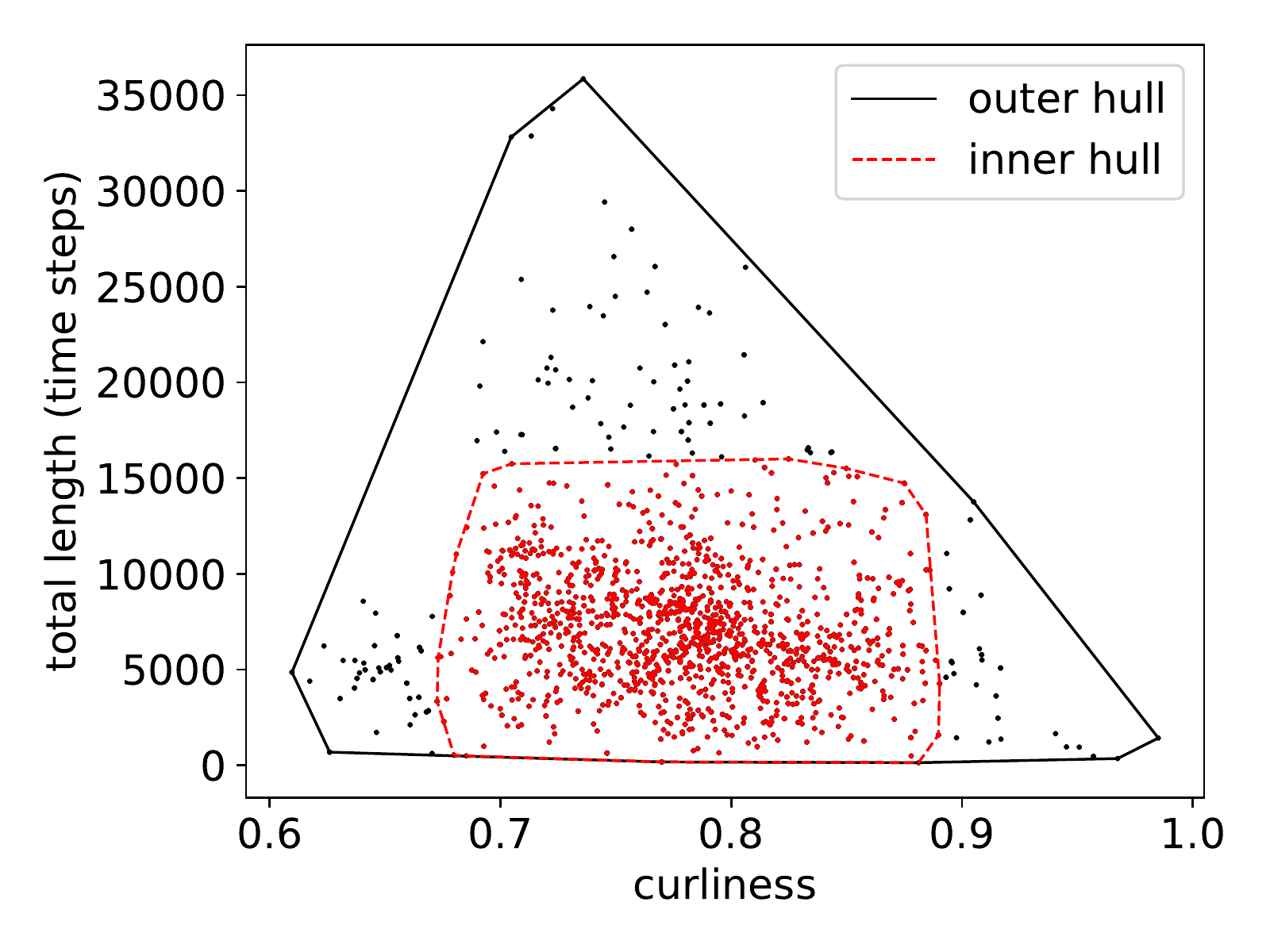}
 		\caption{}
 		\label{fig:landscape_a}
 	\end{subfigure}
 	
 	\begin{subfigure}[b]{0.99\textwidth}
 	 	\centering
 		\includegraphics[scale=0.40]{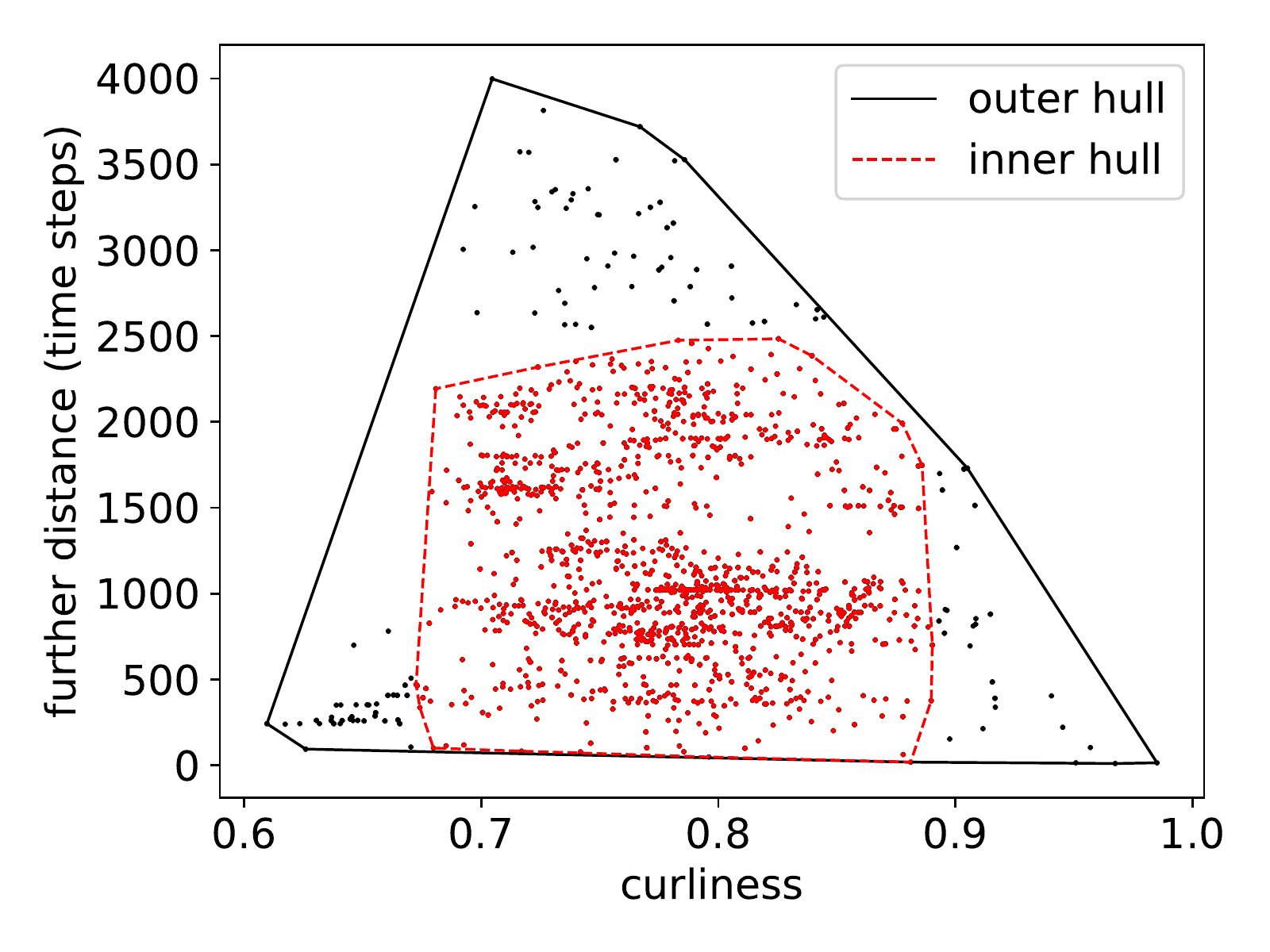}
 		\caption{}
 		\label{fig:landscape_b}
 	\end{subfigure}

 	\begin{subfigure}[b]{0.99\textwidth}
 	    \centering
 		\includegraphics[scale=0.40]{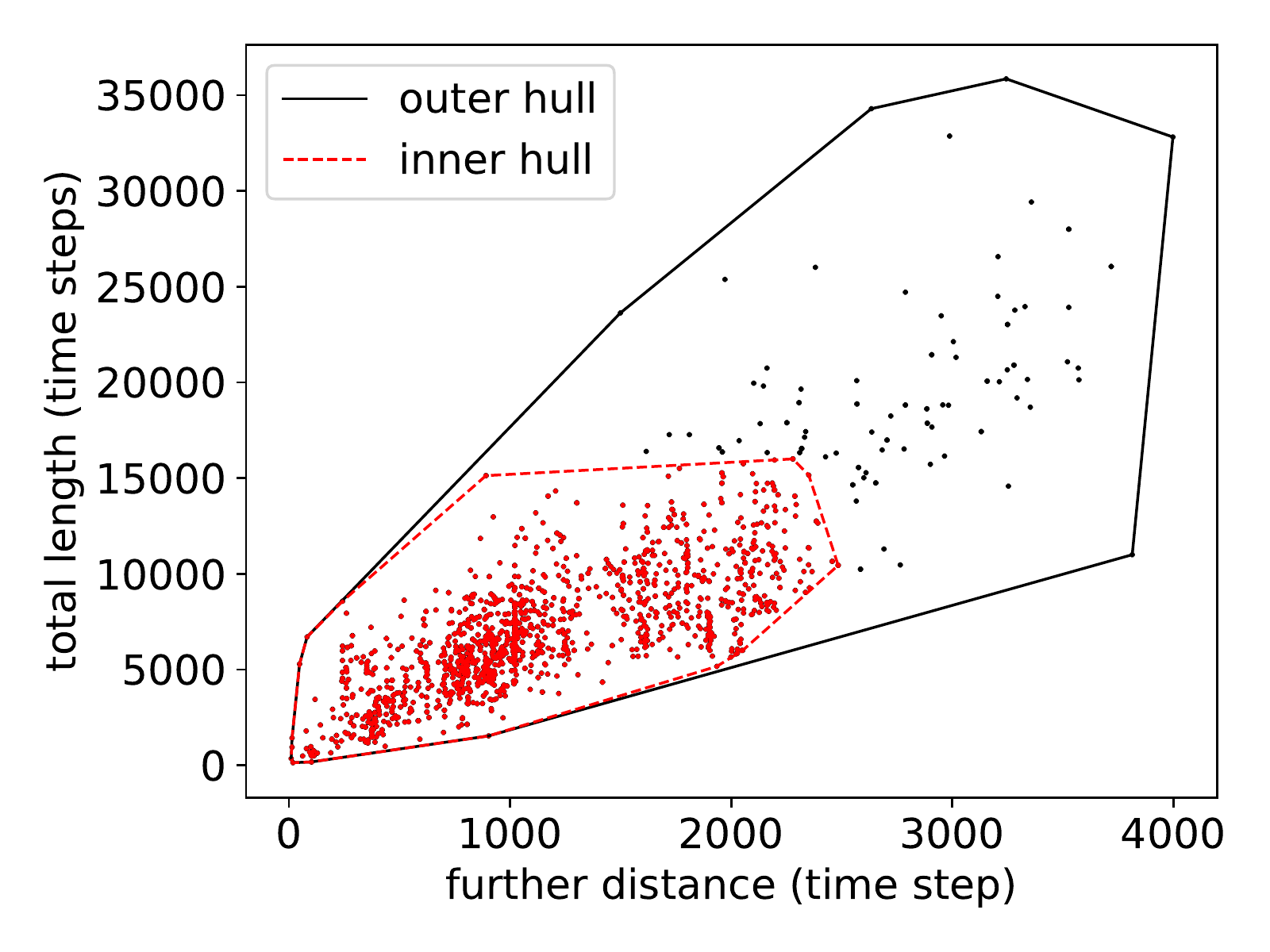}
 		\caption{}
 		\label{fig:landscape_c}
 	\end{subfigure}
 	\caption[]{Visualisation of the features against each other. Figure \ref{fig:landscape_a} represents curliness and total length of the trajectories plotted together. Figure \ref{fig:landscape_b} represents curliness and farthest distance of the trajectories plotted together. Figure \ref{fig:landscape_c} represents further distance and total length of the trajectories plotted together. All the graphs show the outer limit of the landscape and the inner area representing the majority of the points.}
 	\label{img:landscape}

 \end{figure*}

The features are plotted one against each other in order to obtain the landscapes shown in Fig. \ref{img:landscape}.
Every dot in the graphs corresponds to a single trajectory.
The outside borders of the three areas are found using a convex hull algorithm \cite{convex_hull} and they form the legitimate part of the landscape where all the original trajectories lay.
This limit represents the human-likeness border, with values inside this area depicting realistic trajectories, and the outside representing not reasonable trajectories.
A second smaller area has been identified using the Z-score outlier detection system, laying inside the hulls, consisting in the most densely populated area.
The smaller centre areas are the main targets for our methods, representing a plateau of the best feasible solutions.
We will now call this landscape as the\textit{ trajectories reward landscape}, and the metric to define how close a given solution is to reach the plateau as \textit{trajectories reward function} (trf).
\cite{Zonta2020} compares two different trf, with no real differences between the results obtained.
Hence, in this comparison we employ the basic trf:
\begin{equation}\label{eq:features_f}
    f(a)=\left\{\begin{matrix}
        MAX\_TFF & if\:a \in I   \\ 
        d(a, P_I) & if \:a \in O\: \wedge\:a \notin  I   \\ 
        -d(a, P_O)  & otherwise
    \end{matrix}\right.
\end{equation}
with $a$ being the point expressing the two features chosen, $P_I$ the inner area of the trajectories fitness landscape, and $P_O$ the area between the inner hull and the outer one. 
$d$ represents the Euclidean distance between $a$ and the closest side of the polygon.
The idea behind this formulation is to provide a nice gradient for the methods aiming to the plateau, with values in the range $(-\infty, MAX\_TRF]$.
To make the system flexible to accept more features in the future, from the feature representation only a single value is returned:
\begin{equation}\label{eq:feature_representation_total}
k(a,b,c)=z(a) + z(b) + z(c)
\end{equation}
corresponding at the sum of the function $z()$ over the, 3 in our case, combination of features chosen.\\

Other than the features just described, we defined four other to evaluate the final trajectories:
\begin{itemize}
    \item \textbf{no\_overlapping}: the degree in which different trajectories starting from the same starting point overlap between them.
    \item \textbf{directions}: how many directions the trajectories generated from the same starting point have.
    \item \textbf{distance} of the \textbf{ending point} from the \textbf{start}: how far away is the ending point from the starting one. 
    \item \textbf{distance} of the \textbf{middle point} from the \textbf{start}: how far away is the middle point of the trajectory from the starting one.
\end{itemize}
The two distance features are pretty straightforward in their computation. 
No overlapping and direction are two features derived by some characteristics of the trajectories. 
The first one checks how many different cells the trajectories generated posses originating from the same starting point.
Smaller this number is, all the trajectories generated are the same; bigger it becomes, more diverse are the trajectories between them.
The feature ``Direction'' is computed checking the direction of the last point of a trajectory compared to its initial one, measured in cardinal and inter-cardinal points.
The max number for this value is 8, meaning from the same starting point the model can generate versions of the trajectory that all end facing different directions.

These features have not been used to compute the \textit{trf} in order to keep it simple so the landscapes do not explode in complexity. 
This might be a direction for future testing.

\section{Methods}
\label{section:models}
We extended the well-known path-finding algorithm A* in two ways: the first extension includes only the information from the APF in its computation \cite{Zonta2019}, the second one also adds the feature representation explained in Section \ref{section:features_representation} in its computation  \cite{Zonta2020}.
    
A* is an informed search algorithm  \cite{4082128}, meaning that it is formulated in terms of weighted graphs: the algorithm starts from a specific node of a graph and it aims to find a path to a given goal node having the smallest total cost.
The cost is defined, for instance, as the least distance travelled or the shortest time. 
The main loop is done maintaining a tree of paths originating from the starting node and extending those paths one edge at a time until its termination criterion is satisfied.
At each iteration of its main loop, A* needs to determine which paths to extend. 
It does so based on the cost of the path just travelled plus an estimate of the cost required to extend the route to the goal. 
A* selects the path that minimises
$$ f(n)=g(n)+h(n)$$
where \textit{n} is the next node on the path, \textit{g(n)} is the cost of the path from the start node to \textit{n}, and \textit{h(n)} is a heuristic function that estimates the cost of the cheapest path from \textit{n} to the current goal. 
A* terminates when the path it chooses to extend is a path from start to goal or if there are no paths eligible to be continued. 
Normally the heuristic function is defined as problem-specific. 
If the heuristic function is admissible, meaning that it never overestimates the actual cost to get to the goal, A* is guaranteed to return an optimal path from the start to the target.\\

\subsection{Attraction-Based A*}
    The idea of attractiveness of a node is added to the heuristic function.
    The heuristic is then defined as: 
    \begin{equation}\label{eq:astar_zero}
        h(n) = (1 - \Delta) \cdot Q + \Delta \cdot d_{end}
    \end{equation}
    \begin{equation}\label{eq:astar_delta}
        \Delta = \frac{max(0, \alpha - d_{end})}{\alpha + \epsilon}
    \end{equation}
    with $\Delta$ a value representing how important the $d_{end}$ remaining distance to the end over the attraction is, and $Q$ the attraction exerted in the current location computed using Eq. \ref{eq:coulomb}.
    Equation \ref{eq:astar_delta} describes the degree which the distance to the end becomes more important than the attractiveness. 
    The value $\alpha$ defines the limit where the shift happens.
    If the distance to the end is bigger than this value, the equation will return 0, giving importance only to the attractiveness in Eq. \ref{eq:astar_zero}.
    If the distance to the end is smaller than the $\alpha$ value, the $\Delta$ value increases till it reaches 1, which indicates that the heuristic uses only the distance to end as a factor to decide which next node to open.
    The idea behind this formulation resides in how the algorithm works.
    When it is driven by only attraction, in the presence of highly attractive points it can stagnate very close to them, exploring all the nodes and returning a not realistic path.
    With the introduction of distance to travel, the algorithm is forced to go away from these local optima in order to explore the area.\\

\subsection{Feature-Based A*}
    We modified even more the original implementation of the A* algorithm with the addition of feature informations into the heuristic.
    The heuristic is then defined as: 

    \begin{equation}\label{eq:astar_features}
        h(n) = (1 - \Delta) \cdot QR(n)  + \Delta \cdot d_{end}
    \end{equation}
    with $\Delta$ still being Eq. \ref{eq:astar_delta}, $d_{end}$ the remaining distance to the end, $R(n)$ being Eq. \ref{eq:feature_representation_total} with the features computed from the trajectory generated so far, and $Q$ the attraction exerted in the current location computed using Eq. \ref{eq:coulomb}.
    This heuristic is slower compared to the previous one, since the algorithm has to remember all the trajectories tested so far for the computation of Eq. \ref{eq:feature_representation_total}.
    The idea behind this modification is to use more information from the real trajectories to generate new ones.

\section{Dataset}
\label{section:dataset}
\begin{figure}[t]
 	\centering 
 	\includegraphics[scale=0.58]{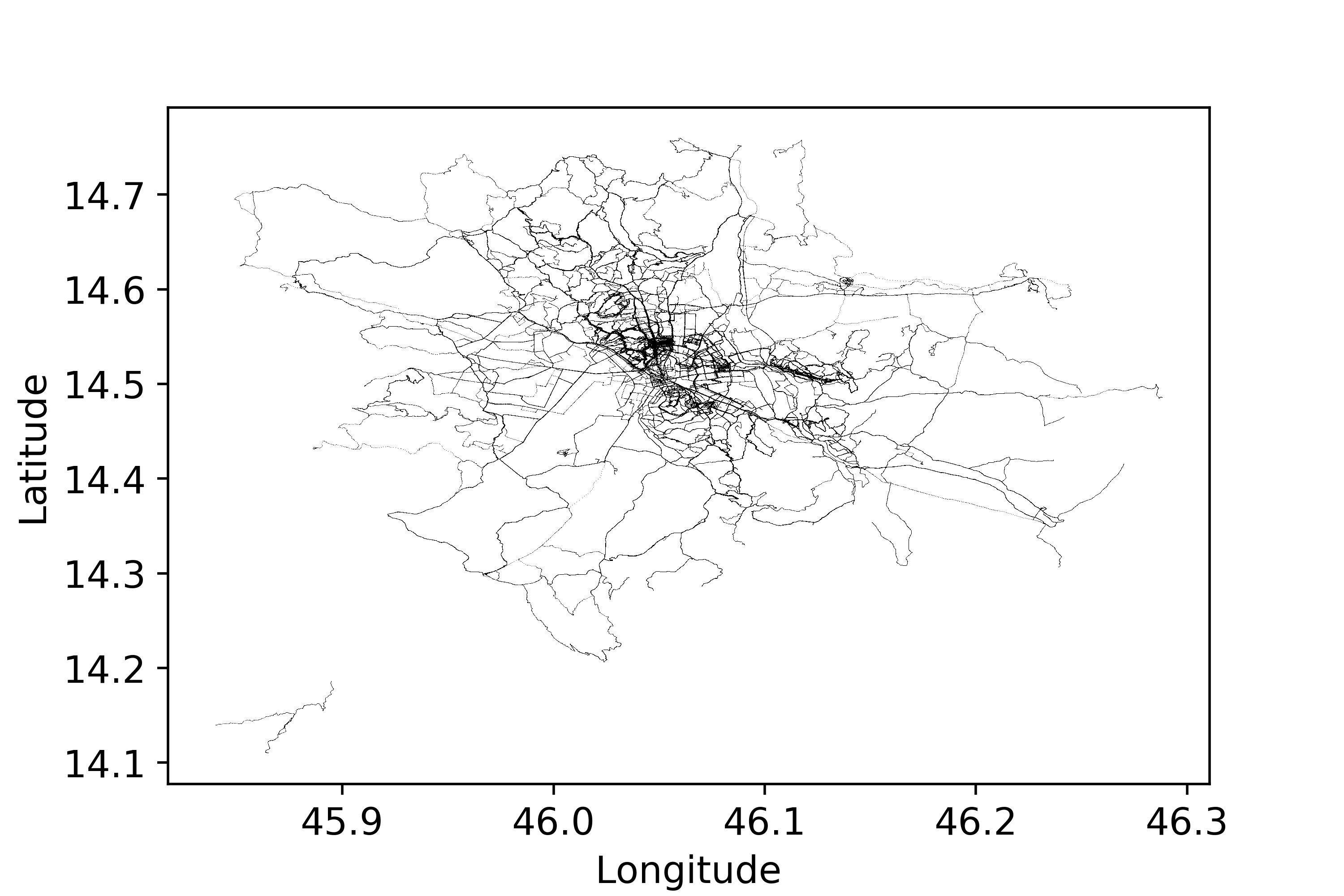} 
 	\caption{Visualisation of all the trajectories available in the dataset.}
 	\label{img:map}
 \end{figure}

% Description of the dataset with some info
As a dataset containing real trajectories, we used  \cite{Connect2015} which contains data produced by nine cyclists for a total of 3089 trajectories. 
Data were directly exported from their Strava or Garmin Connect accounts, consisting of GPS locations, elevation, duration, distance, average and maximal heart rate, and for some workouts also some information is obtained from power meters.
Trajectories were analysed to remove outliers to keep the APF dimensions contained.
The final area resulting from the preprocessing correspond to $4212km^2$.
In the data, round-trips are also present in small amount, with only 205 trajectories ending less than $25$ meters from the start.
Table \ref{table:datasetSettings} shows some statistics about the dataset, and \ref{img:map} shows a visualisation of the distribution of the final trajectories.
The city centre of Ljubljana is visible given the high density of trajectories located there.
The real trajectories are adapted to the APF representation following an iterative process.
Two consecutive points of the real trajectories are matched in the APF matrix and connected thanks to a standard A* algorithm.
Iterating this process over all the points in a trajectory, we obtain its APF matrix representation that can be used for low lever comparison between paths.

\begin{table}[t]
	\centering
	\caption{Dataset Details.}
	\label{table:datasetSettings}
	\begin{tabular}{lll}
		\toprule
		Features                                          & Mean             & Std  \\
		\midrule
		Number of time-steps                     & 2165.39         &  1955.24                  \\
		Length of trajectories ($m$)                & 40238.15       & 22315.4                   \\
		Speed ($\frac{m}{s}$)                       & 5.8                & 11.0                          \\
		Space between time-steps ($m$)    & 18.6              & 35.9                         \\
		\bottomrule
	\end{tabular}
\end{table}

\section{Experimental Setup}
\label{section:experimental_setup}
All the A*\footnote{github.com/AleZonta/TrajectoriesAstar} experiments are repeated with 30 different seeds, generating trajectories from 15 different starting point with 30 distinctive environmental multipliers and $\alpha$ values from Eq. \ref{eq:astar_zero} fixed to 50.
The algorithms are executed with generative max lengths of the trajectories in the range 500-10000, with step 1000.

Both algorithms return a path, i.e. a list of APF cells connected by a border or a corner.
Paths follow the routing system of the area, but since they are a simple list of contiguous cells, they cannot be yet defined as trajectories.
To transform the path into a trajectory, only the cells distant from each other by value drawn from a \textit{lognormal distribution} are kept.
The distribution is fit using the information contained in the training set.

To check the efficiency of the two methods, the experiments are run in an isolated environment where the execution time is recorded.
Every experiment generated the same amount of trajectories, and the programs are equipped with the multiprocessing paradigm to speed up the computation.
This translates into the idea that every trajectory is computed in its own single core.
To check the efficacy of the two algorithms, the average \textit{trf} of the methods and the real trajectories is analysed together with the other feature described in Section \ref{section:features_representation}.
The two algorithms are compared with the real trajectories modified to respect the total length of the generated trajectories.
The Wilcoxon signed-rank test for differences in the means (averaged over all runs) is employed to statistically analyse the data.

The next experiments performed also save the A*  algorithm information on a file, to enable an in-depth analysis of the search algorithm on the map. 
This procedure did not occur with the previous experiment in order to reduce any overhead caused by I/O procedure for saving the data.
The number of trajectories generated increases to 50 different starting points instead of the initial 15.

In order to test the variation of the $\alpha$ values from Eq. \ref{eq:astar_zero}, all the value from 1 to 100 with step 5 are tested.
The algorithm automatically converts this value as the percentage of the total distance to assign to $\alpha$.
This process produces 21 different combinations of the parameter.
To simplify the analysis, we selected 5000 meters as maximum length allowed on the generated trajectories for these experiments, given the great trade-off between how long the trajectories are and the time to create them.
The environmental multipliers are fixed to the same attractive value for all the tags for this set of experiments.
For this experiment, also the raw details of the two A* algorithm are highlighted for better comprehension of the hyperparameter effects\footnote{more graphs are visible at \url{https://github.com/AleZonta/AlphaParameter}}.

In order to test the environmental control over the final trajectories, we define multipliers which indicate how attractive a family of tags is.
The raw description of a multiplier is a vector $<B, A, N, O, Sh, Sp>$ with every letter correspond to the initial of the related tag. 
As a consequence of limiting the value of the charge from 1 to 100, the multipliers are obtained by all the permutation without repetition of the basic vector $<1, 25, 10, 75, 50, 100>$.
This permutation of values produces 720 different combinations of multipliers.
To simplify the analysis, we selected 5000 meters as maximum length allowed on the generated trajectories for these experiments, given the great trade-off between long the trajectories are and the time to create them.
The $\alpha$ value is fixed to half attraction and half distance to the end for this experiments, specifically $\alpha = 50$.

To quantify the effect of the different hyper-parameters on the generated trajectories, we used the Dynamic Time Warping (DTW) \cite{Berndt} measure to assess the similarity between trajectories.
This measure defines the variations the hyper-parameters are able to achieve on the trajectories generated from the same starting point. 
For every starting point, the similarity between the different variations of trajectories is computed and then ordered and plotted in a graph divided into the percentage of variation.
This measurement can help us understand how the algorithm behaves with the inclusion of more diverse outcomes in the pool of analysed trajectories.
The Wilcoxon signed-rank test for differences in the means (averaged over all runs) is employed to statistically analyse the data.

\section{Results}
\label{section:results}
This paper aims to conduct an in-depth analysis of the two approaches following several key points regarding their efficiency, their efficacy and the power of their hyper-parameters to condition the final trajectories.
The two preeminent hyperparameters tested are the value $\alpha$ from Eq. \ref{eq:astar_zero} that controls the preference between total distance or attractiveness of the environment and the multipliers that manage the sensitivity to environmental features.
Given this clear division of points, we illustrate the result of each of them in the following paragraphs.

\subsection{Efficiency}
    \begin{figure}[!ht]
        \centering
 		\includegraphics[scale=0.60]{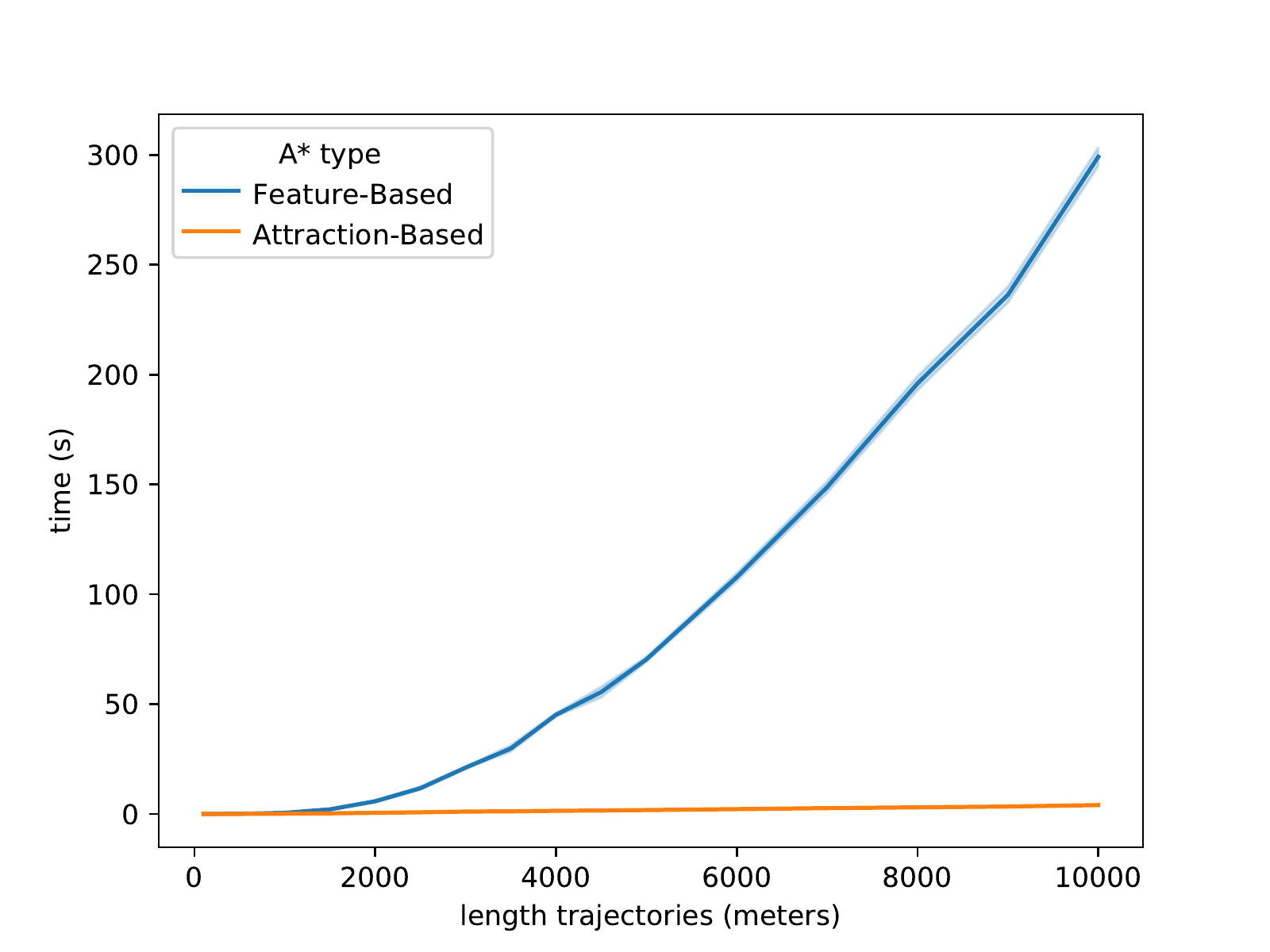}
 		\caption{Time efficiency comparison between the two version of A*. The time in seconds is plotted over different length of generated trajectories.}
 		\label{img:result:time}
    \end{figure}
    \begin{figure}[!h]
        \centering
     		\includegraphics[scale=0.45]{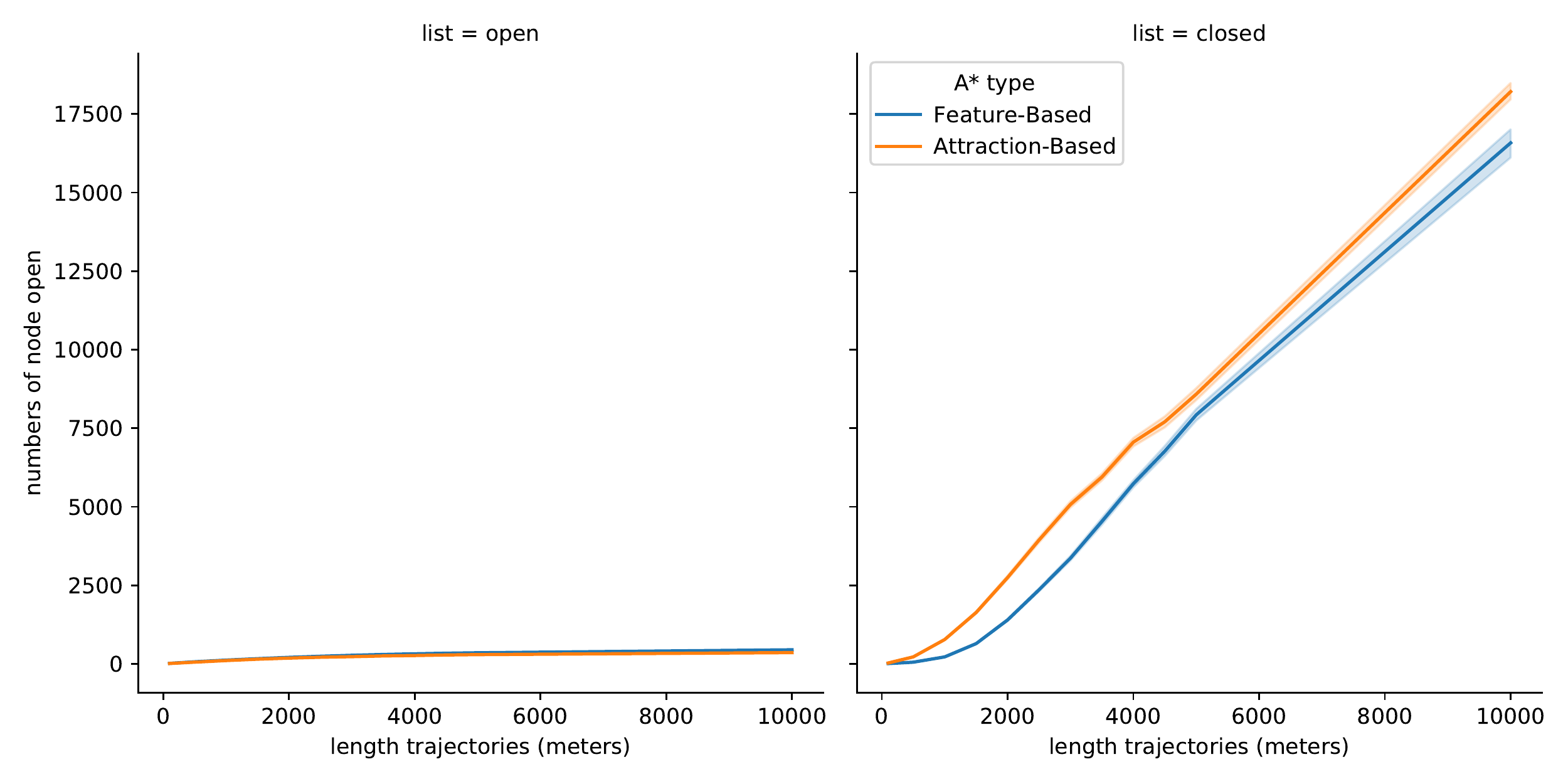}
     		\caption{Comparison between the number of open and closed nodes by the two versions of A*. The raw number of nodes is plotted over different length of generated trajectories.}
     		\label{img:result:nodes}
     \end{figure}
    
    Figure \ref{img:result:time} shows a comparison of the time efficiency between the two versions of A*.
    The comparison shows the length of the trajectories generated by the A* compared to the time in seconds taken by the algorithms to return them.
    Clearly visible is the exponential growth of the Feature-Based A*, compared to the linear growth of the Attraction-Based A*.
    
    Figure \ref{img:result:nodes} shows the comparison of the sampling efficiency between the open and closed nodes of the two algorithms during the generation of trajectories of different lengths.
    Meanwhile the number of open nodes remains constant between the two A* types and the length of the trajectories to generate, the number of closed nodes grows linearly with the increase of the length of the trajectories.
    Interesting to notice the Feature-Based A* always closes fewer nodes than the Attraction-Based A*, even though the execution time is way higher.
    This discrepancy can be explained considering how the two versions work. 
    The only difference between the two algorithms is the addition of the computation of the features, which helps to reduces the number of nodes to visit before finding the trajectories at the expense of time.

\subsection{Efficacy}

    \begin{figure}[!h]
        \begin{subfigure}[b]{0.49\textwidth}
     	 	\centering
     		\includegraphics[scale=0.4]{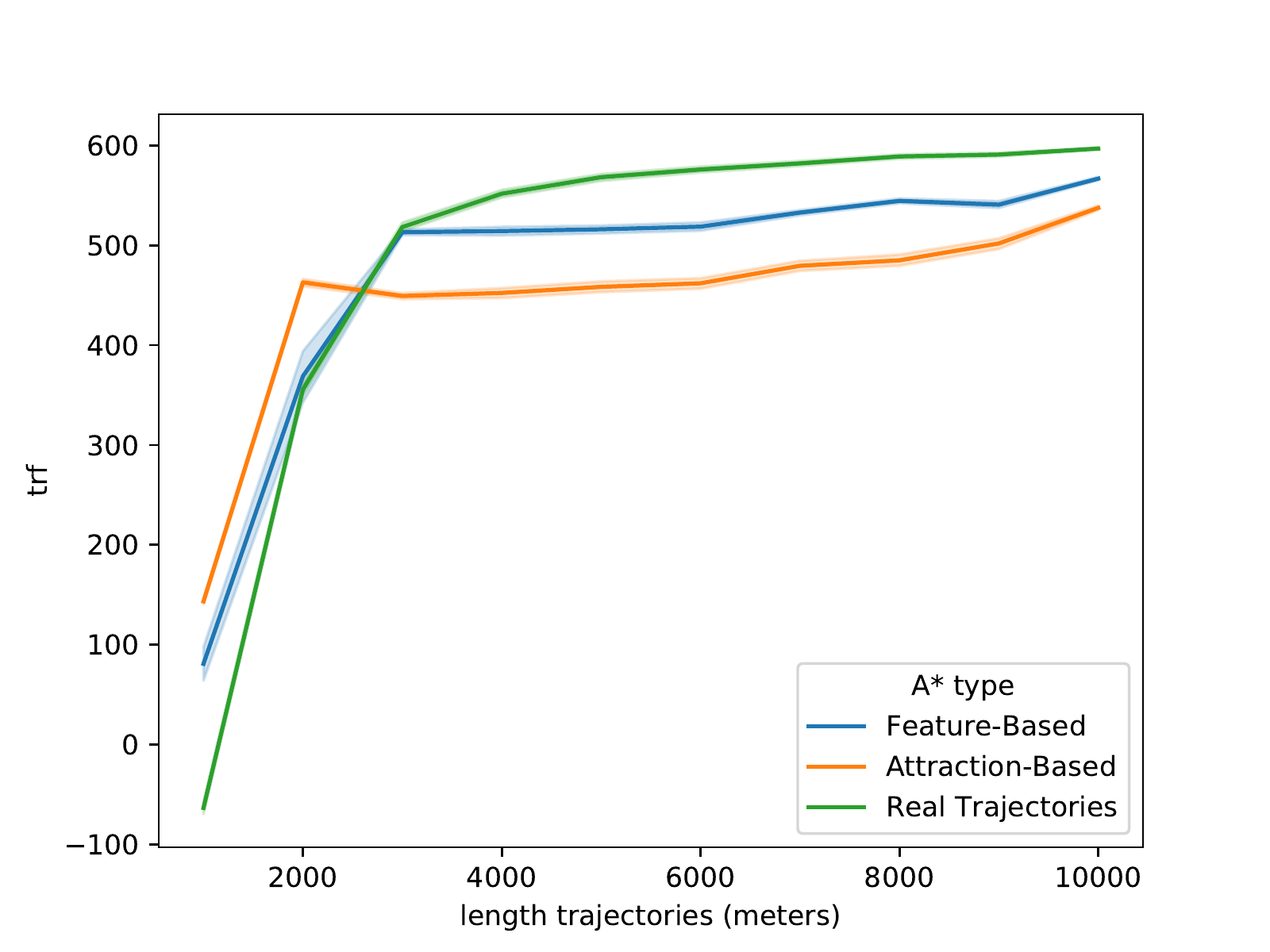}
     		\caption{}
     		\label{img:result:fitness}
     	\end{subfigure}
     	\begin{subfigure}[b]{0.49\textwidth}
     	 	\centering
     		\includegraphics[scale=0.4]{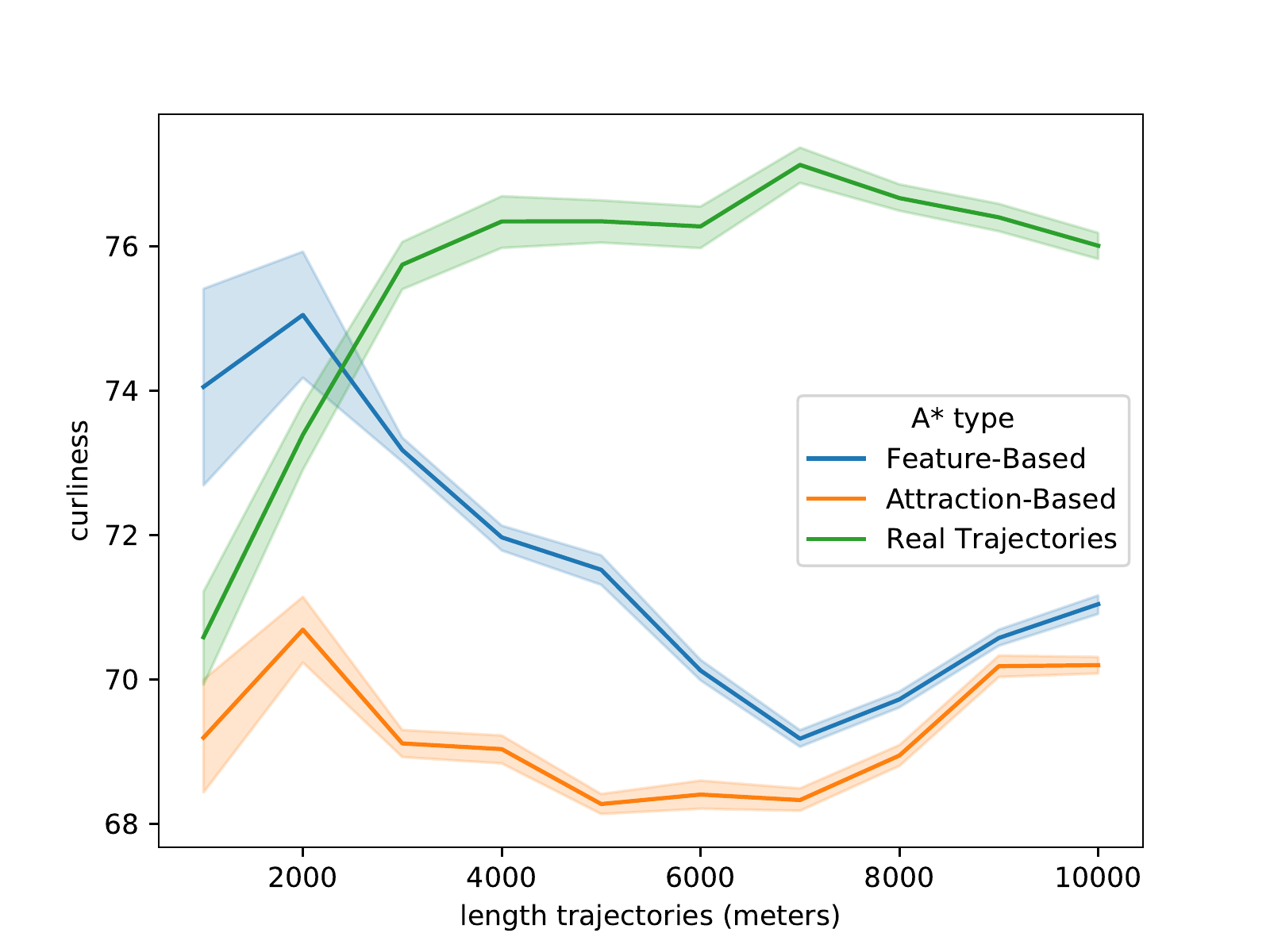}
     		\caption{}
     		\label{img:result:curliness}
     	\end{subfigure}
         
         \begin{subfigure}[b]{0.49\textwidth}
     	 	\centering
     		\includegraphics[scale=0.4]{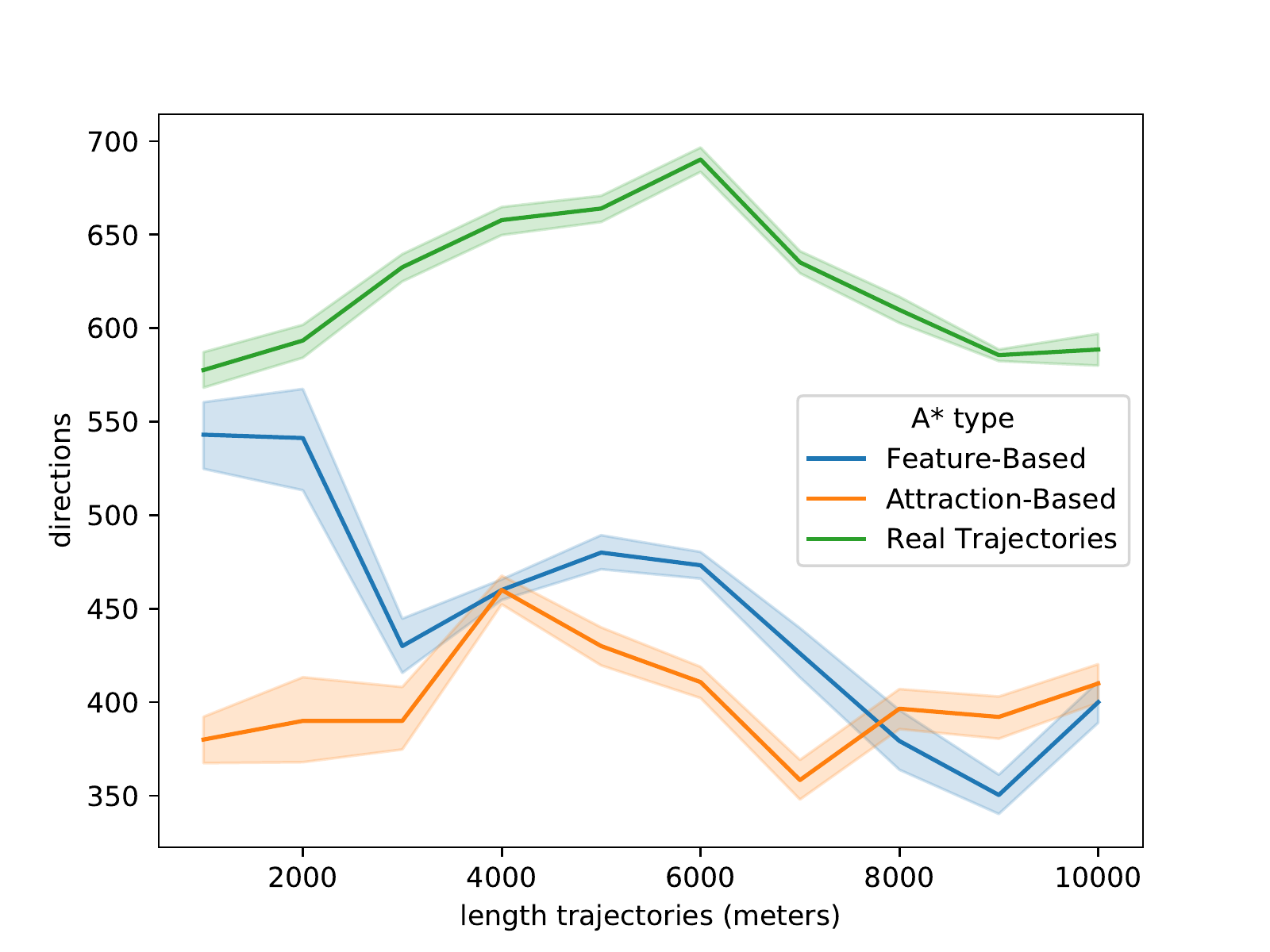}
     		\caption{}
     		\label{img:result:directions}
     	\end{subfigure}
     	\begin{subfigure}[b]{0.49\textwidth}
     	 	\centering
     		\includegraphics[scale=0.4]{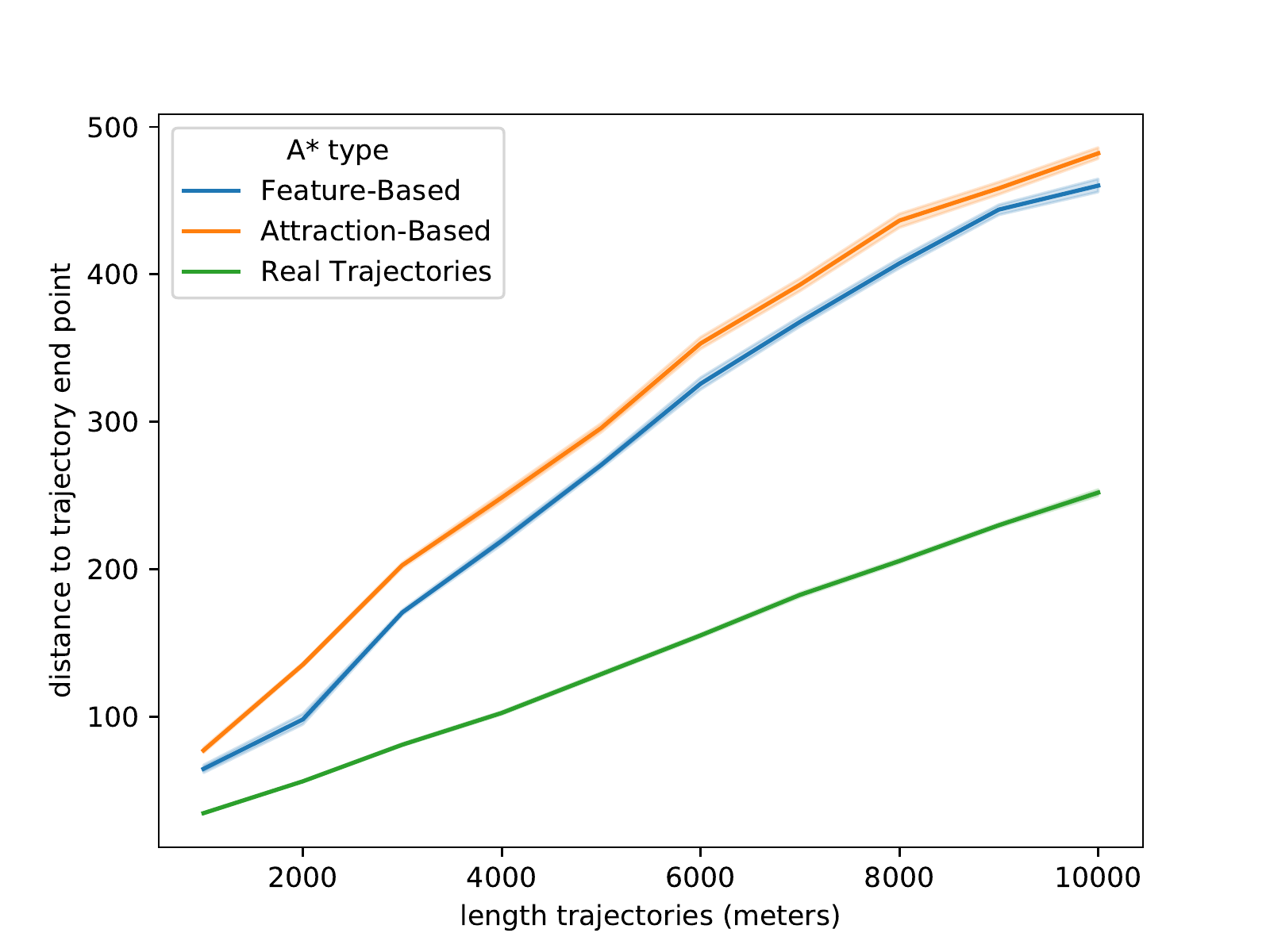}
     		\caption{}
     		\label{img:result:distance_to_end}
     	\end{subfigure}
     	\caption{Comparison between the two A* algorithm and the real trajectories using the features described in Section \ref{section:features_representation}. All the graphs show a feature over different lengths of generated trajectories, specifically Fig. \ref{img:result:fitness} shows the \textit{trf}, Fig. \ref{img:result:curliness} the curliness, Fig. \ref{img:result:directions} the amount of directions achieved, and Fig. \ref{img:result:distance_to_end} the distance to the end point of the trajectories.}
     	\label{img:result:result_features_a}
     \end{figure}
     \begin{figure}[!h]
     	\begin{subfigure}[b]{0.49\textwidth}
     	 	\centering
     		\includegraphics[scale=0.4]{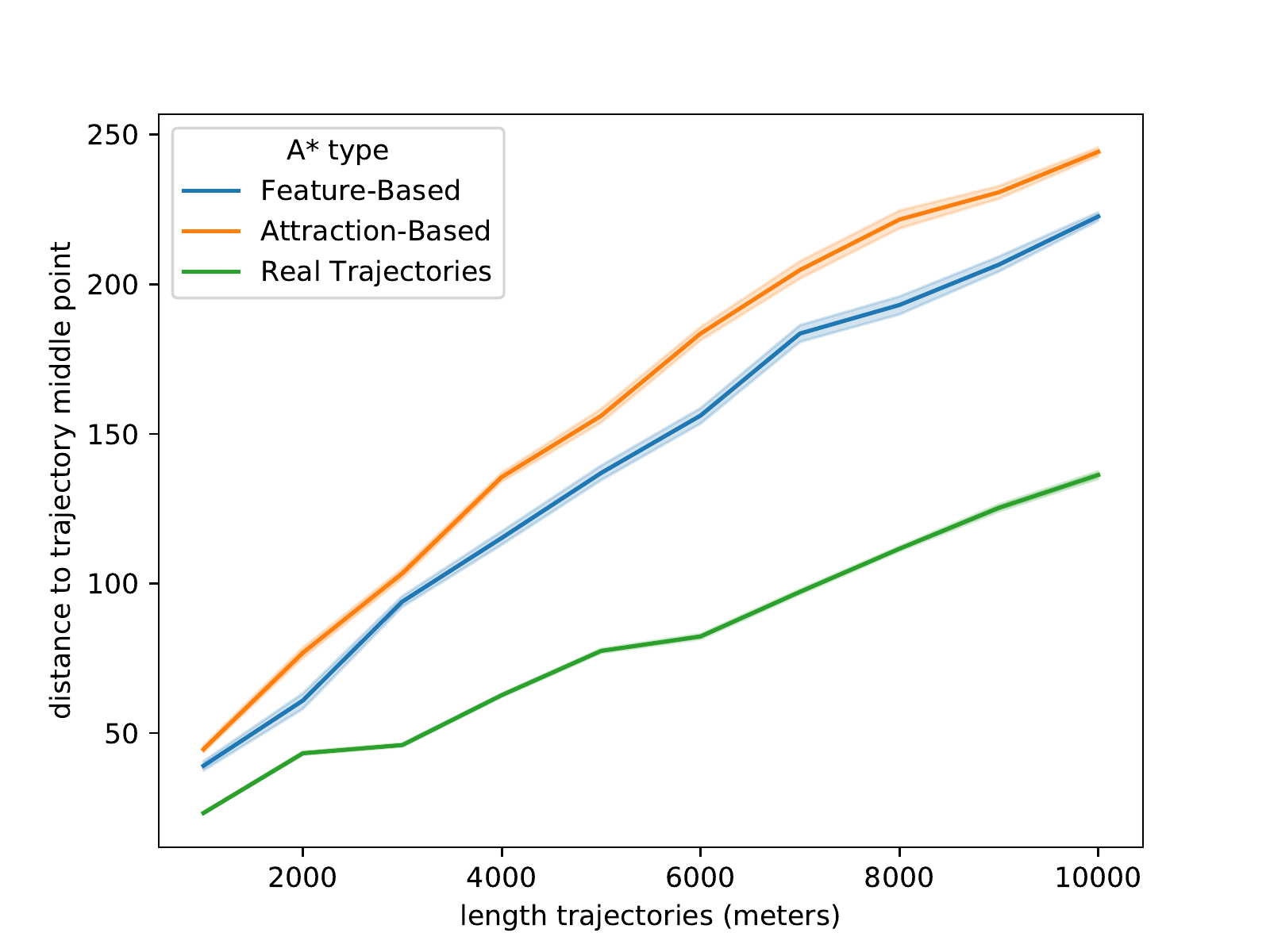}
     		\caption{}
     		\label{img:result:distance_to_middle_point}
     	\end{subfigure}
     	\begin{subfigure}[b]{0.49\textwidth}
     	 	\centering
     		\includegraphics[scale=0.4]{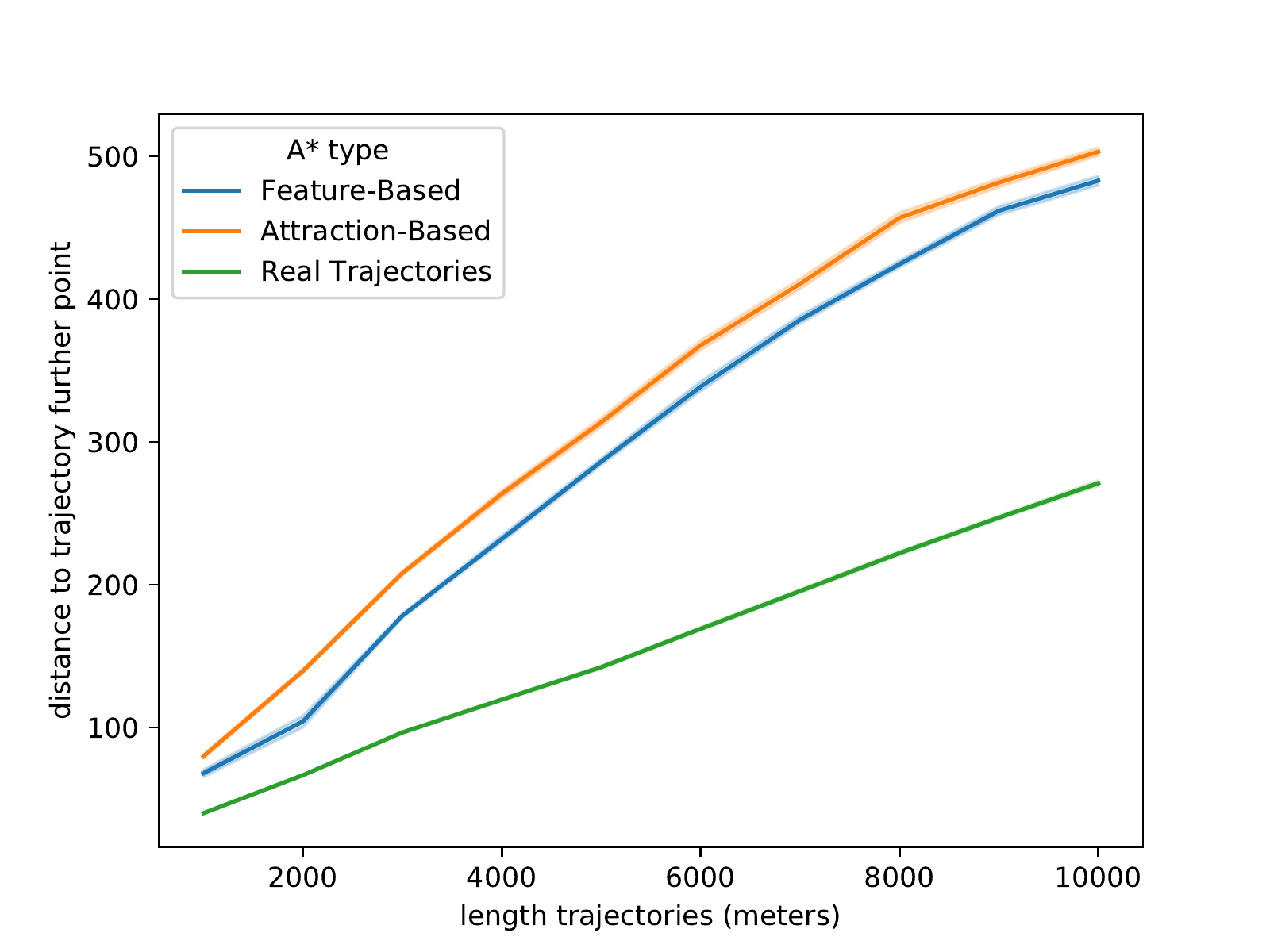}
     		\caption{}
     		\label{img:result:_distance_to_further_point}
     	\end{subfigure}
     	
     	\begin{subfigure}[b]{0.99\textwidth}
     	 	\centering
     		\includegraphics[scale=0.4]{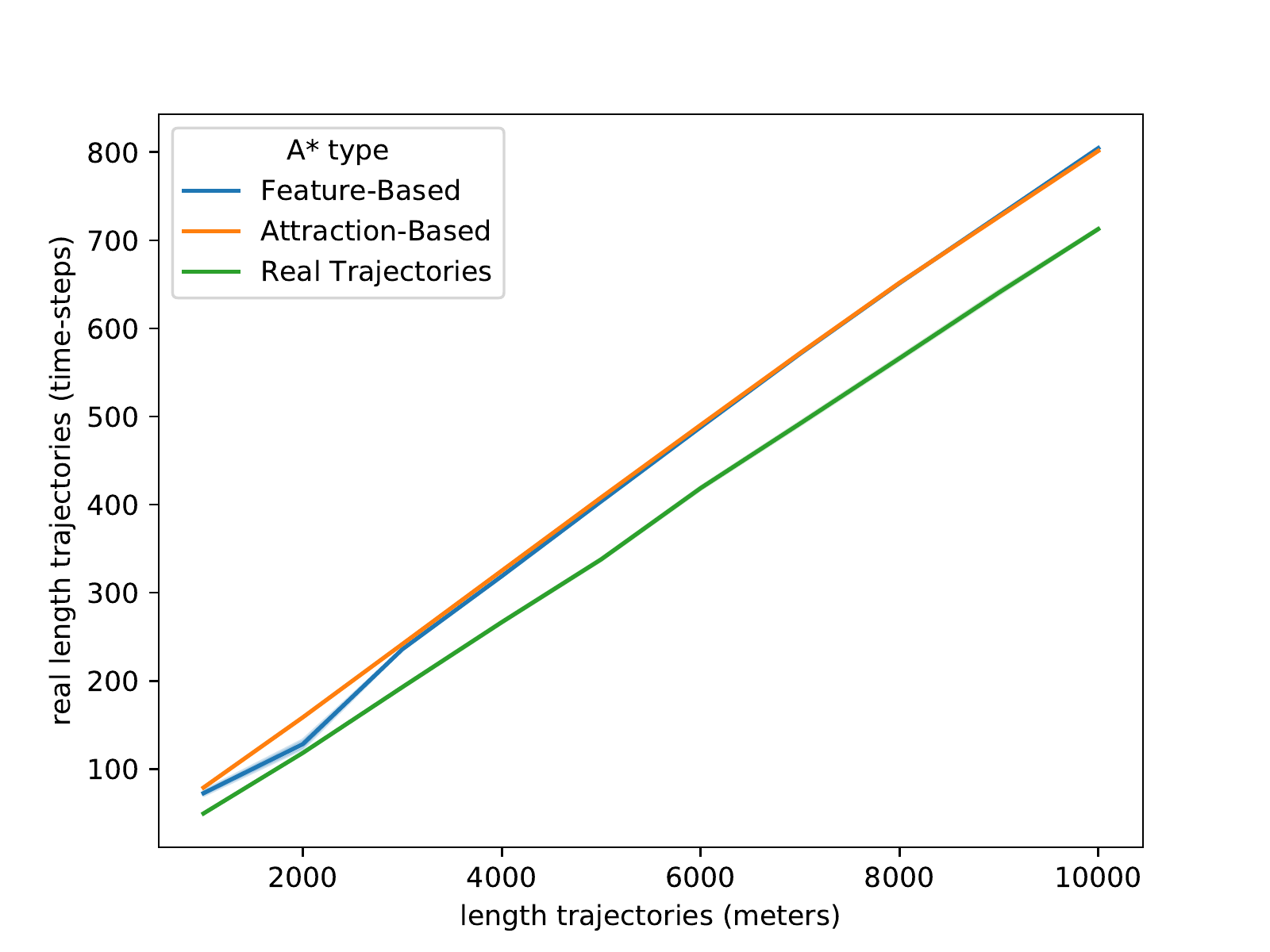}
     		\caption{}
     		\label{img:result:total_length}
     	\end{subfigure}
     	\caption{Comparison between the two A* algorithms and the real trajectories using the features described in Section \ref{section:features_representation}. All the graphs show a feature over different lengths of generated trajectories, specifically Fig. \ref{img:result:distance_to_middle_point} shows the distance to the middle point of the trajectories, Fig. \ref{img:result:_distance_to_further_point} the distance to the farthest point from the starting point of the trajectories, and Fig. \ref{img:result:total_length} the total length measured in time-steps.}
     	\label{img:result:result_features_b}
  \end{figure}
    
    Figures \ref{img:result:result_features_a} and  \ref{img:result:result_features_b} present a comparison between the two A* algorithms and the real trajectories using the features representation applied to the generated and real trajectories.
    All the graphs represent the features analysed computed over different lengths of generated trajectories.
    Figure \ref{img:result:fitness} represent the \textit{trf} with all the three models analysed behaving similarly, with very fast growth at the beginning and a very slow on after $~3000$ meters long trajectories.
    Interestingly, before this cutoff value, the Attraction-Based A*  algorithm produces better scores than the real trajectories and the Feature-Based A*  algorithm but then it degrades to be least performing model.
    Between $2000$ and $3000$ meters, the performances of the Feature-Based A*  algorithm coincide with the one from the real trajectories.
    After this point, a decrease of performances is visible, where the model never reaches again the values of the real trajectories.
    With the increase of the total length of trajectories to generate, the two A* algorithms seem to slowly increase the performances, which end with an upward trending.
    The Feature-Based A*  algorithm achieves slightly better performances than the Attraction-Based A* algorithm, going closer to the values from the real trajectories.
    The injection of information about the real trajectories into the model is proven successful only for very short trajectories, as opposed to the effect with longer ones, where it just helps a little bit but never matching the target values.
    
    Figure \ref{img:result:curliness} presents the curliness feature, with the two models never able to reach the same values of the real trajectories.
    Also, in this case, the Feature-Based A*  algorithm reaches values closer to the real trajectories compared to the Attraction-Based A* algorithm.
    
    Figure \ref{img:result:directions} depicts the average amount of directions the trajectories faces.
    The two models behave similarly producing always less direction that the real trajectories.
    Seems like the A*  algorithms when they produce longer trajectories they end up with the ending point facing always the same directions, reducing the diversity of the trajectories.
    The two methods clearly luck into the ability to produce the number of directions present in the real dataset.
    This detail is also visible in Figures \ref{fig:results:normal_astar_alpha}, \ref{fig:results:feature_astar_alpha}, \ref{fig:results:normal_astar_behaviour}, and \ref{fig:results:feature_astar_behaviour}.
    
    Figures \ref{img:result:distance_to_end}, \ref{img:result:distance_to_middle_point}, and \ref{img:result:_distance_to_further_point} shows the distance to the end point, middle point, and further point of the trajectories from the starting point. 
    All the three cases show the same pattern, with the two A* methods increasing distances way faster than the real trajectories over longer trajectories generated.
    In all the three cases, the Feature-Based A* algorithm has closer value to the real trajectories, even though still at a sensible distance from them.
    The real trajectories are reduced to match the distance of the generated ones for a fair comparison.
    Given the fact the real one is way longer than the generated one, this adaptation causes the line on the graphs to not show any variation in it.
    
    Figure \ref{img:result:total_length} shows the comparison of the time-steps of the real and generated trajectories over the length of the trajectories generated.
    The two A* algorithms produces exactly the same amount of time-steps trajectories, always a little bit more than the real trajectories.
    This result is exciting, considering the data showed on Fig. \ref{img:result:nodes}, where the Feature-Based A*  algorithm is visiting fewer nodes than the Attraction-Based one.
    This development means the Feature-Based A*  algorithm is more efficient in finding the trajectories compared to the Attraction-Based one.
    
    To understand why the Feature-Based A*  algorithm fails in reaching the real trajectories performances, analysing these graphs together with Fig. \ref{img:landscape} can give us the full picture.
    Even though the average of the curliness from the two methods is not the same as the one from the real trajectories, they are always inside the plateau showed in Fig. \ref{img:landscape}. 
    On the other hand, the distances are always larger than the one from the real trajectories, decreasing the total \textit{trf} and therefore failing to reach the target one.
    The Attraction-Based A*  algorithm does not possess this information during the generation of the trajectories, hence its lower performances on basically all the feature we used to compare the two algorithms.
    
    Statistical test comparing the curves in Figures \ref{img:result:result_features_a} and  \ref{img:result:result_features_b} are shown in Tables \ref{table:statistical_test_a}, \ref{table:statistical_test_b}, \ref{table:statistical_test_c}, and \ref{table:statistical_test_d}
    
\begin{table}[h!]
	\centering
	\caption{Wilcoxon test ($p$-values) per every length of generated trajectories of the \textit{trf} and curliness feature. Significant differences ($p < 0.05$) are highlighted. $F_A$ = Feature-Based A*, $A_A$ = Attraction-Based A*, $R$ = Real Trajectories}
	\label{table:statistical_test_a}
\begin{tabular}{ll@{\hskip 0.1in}l@{\hskip 0.1in}l@{\hskip 0.1in}l@{\hskip 0.1in}l@{\hskip 0.1in}l}
		\toprule

      & \multicolumn{3}{c}{\textit{trf}}                             & \multicolumn{3}{c}{curliness}   \\
      & $F_A\: vs\: A_A$    & $F_A\: vs\: A_A$      & $A_A\: vs\: R$      & $F_A\: vs\: A_A$  & $F_A\: vs\: R$ &      $A_A\: vs\: R$     \\
      		\midrule
    & \multicolumn{1}{c}{$p <$ }   &  \multicolumn{1}{c}{$p <$ }    &  \multicolumn{1}{c}{$p <$ }    & \multicolumn{1}{c}{$p <$ }  &  \multicolumn{1}{c}{$p <$ }  &     \multicolumn{1}{c} {$p <$ }   \\
          		\midrule
1000  & \textbf{4.1e-09} & \textbf{2.6e-23} & \textbf{1.7e-39} & \textbf{5.5e-16} & \textbf{0.0001}  & \textbf{0.0078}  \\
2000  & 0.9445           & 0.4884           & \textbf{1.8e-37} & \textbf{3.2e-27} & \textbf{0.0156}  & \textbf{7.5e-15}  \\
3000  & \textbf{3.2e-54} & \textbf{0.0352}  & \textbf{2.0e-45} & \textbf{3.2e-54} & \textbf{1.0e-23} & \textbf{1.9e-50} \\
4000  & \textbf{5.9e-51} & \textbf{1.3e-21} & \textbf{5.9e-47} & \textbf{6.0e-51} & \textbf{1.8e-32} & \textbf{3.0e-47}  \\
5000  & \textbf{1.1e-47} & \textbf{4.4e-29} & \textbf{1.9e-45} & \textbf{1.1e-47} & \textbf{3.7e-32} & \textbf{1.9e-45}  \\
6000  & \textbf{5.3e-46} & \textbf{6.2e-46} & \textbf{4.9e-46} & \textbf{3.4e-45} & \textbf{4.9e-46} & \textbf{4.9e-46}  \\
7000  & \textbf{1.5e-30} & \textbf{3.2e-45} & \textbf{3.2e-45} & \textbf{1.6e-15} & \textbf{3.2e-45} & \textbf{3.2e-45} \\
8000  & \textbf{5.0e-42} & \textbf{6.6e-44} & \textbf{6.6e-44} & \textbf{1.0e-07} & \textbf{6.6e-44} & \textbf{6.6e-44} \\
9000  & \textbf{7.2e-42} & \textbf{3.1e-44} & \textbf{2.2e-45} & \textbf{4.8e-15} & \textbf{2.2e-45} & \textbf{2.2e-45}\\
10000 & \textbf{5.9e-51} & \textbf{9.9e-50} & \textbf{6.0e-51} & \textbf{1.5e-45} & \textbf{1.1e-50} & \textbf{6.1e-51} \\
		\bottomrule
\end{tabular}
\end{table}

\begin{table}[h!]
	\centering
	\caption{Wilcoxon test ($p$-values) per every length of generated trajectories of the direction and no overlapping feature. Significant differences ($p < 0.05$) are highlighted. $F_A$ = Feature-Based A*, $A_A$ = Attraction-Based A*, $R$ = Real Trajectories}
	\label{table:statistical_test_b}
\begin{tabular}{ll@{\hskip 0.1in}l@{\hskip 0.1in}l@{\hskip 0.1in}l@{\hskip 0.1in}l@{\hskip 0.1in}l}
		\toprule

      & \multicolumn{3}{c}{direction}                             & \multicolumn{3}{c}{no overlapping}   \\
      & $F_A\: vs\: A_A$    & $F_A\: vs\: A_A$      & $A_A\: vs\: R$      & $F_A\: vs\: A_A$  & $F_A\: vs\: R$ &      $A_A\: vs\: R$     \\
      		\midrule
    & \multicolumn{1}{c}{$p <$ }   &  \multicolumn{1}{c}{$p <$ }    &  \multicolumn{1}{c}{$p <$ }    & \multicolumn{1}{c}{$p <$ }  &  \multicolumn{1}{c}{$p <$ }  &     \multicolumn{1}{c} {$p <$ }   \\
          		\midrule
1000  & \textbf{1.5e-34}                 & \textbf{1.4e-08} & \textbf{4.6e-36} & 0.2795           & \textbf{1.2e-32} & \textbf{2.5e-31} \\
2000  & \textbf{1.1e-30}                 & \textbf{2.8e-05} & \textbf{7.2e-28} & \textbf{0.0086}  & \textbf{0.0182}  & \textbf{0.0330}  \\
3000  & \textbf{2.3e-06}                 & \textbf{3.0e-46} & \textbf{6.0e-47} & \textbf{2.5e-37} & \textbf{6.4e-51} & \textbf{2.5e-47} \\
4000  & 1.0000                           & \textbf{6.8e-50} & \textbf{1.5e-49} & \textbf{6.1e-51} & \textbf{6.5e-50} & \textbf{1.8e-47} \\
5000  & \textbf{1.7e-18}                 & \textbf{1.1e-40} & \textbf{9.0e-44} & \textbf{1.6e-34} & \textbf{2.9e-47} & \textbf{1.1e-46} \\
6000  & \textbf{5.9e-31}                 & \textbf{1.7e-46} & \textbf{1.7e-46} & \textbf{3.4e-21} & \textbf{2.1e-45} & \textbf{1.7e-44} \\
7000  & \textbf{6.7e-10}                 & \textbf{2.8e-38} & \textbf{1.8e-43} & \textbf{0.0006}  & \textbf{2.8e-44} & \textbf{6.0e-44} \\
8000  & \textbf{0.0009}                  & \textbf{1.3e-37} & \textbf{1.0e-42} & \textbf{2.2e-06} & \textbf{2.0e-43} & \textbf{2.1e-43} \\
9000  & \textbf{1.4e-09}                 & \textbf{3.3e-45} & \textbf{6.5e-43} & \textbf{3.0e-16} & \textbf{1.0e-44} & \textbf{4.8e-45} \\
10000 & \textbf{0.0180}                  & \textbf{1.2e-39} & \textbf{3.7e-38} & \textbf{9.0e-16} & \textbf{8.1e-51} & \textbf{6.9e-51} \\
		\bottomrule
\vspace{2cm}

\end{tabular}
\end{table}

\begin{table}[h!]
	\centering
	\caption{Wilcoxon test ($p$-values) per every length of generated trajectories of the distance to trajectory end point and distance to trajectory middle point feature. Significant differences ($p < 0.05$) are highlighted. $F_A$ = Feature-Based A*, $A_A$ = Attraction-Based A*, $R$ = Real Trajectories, f\_d\_t\_p: distance of the farthest point from the start, d\_t\_m\_p: distance to trajectory middle point}
	\label{table:statistical_test_c}
\begin{tabular}{ll@{\hskip 0.1in}l@{\hskip 0.1in}l@{\hskip 0.1in}l@{\hskip 0.1in}l@{\hskip 0.1in}l}
		\toprule

      & \multicolumn{3}{c}{f\_d\_t\_p}                             & \multicolumn{3}{c}{d\_t\_m\_p}   \\
      & $F_A\: vs\: A_A$    & $F_A\: vs\: A_A$      & $A_A\: vs\: R$      & $F_A\: vs\: A_A$  & $F_A\: vs\: R$ &      $A_A\: vs\: R$     \\
      		\midrule
    & \multicolumn{1}{c}{$p <$ }   &  \multicolumn{1}{c}{$p <$ }    &  \multicolumn{1}{c}{$p <$ }    & \multicolumn{1}{c}{$p <$ }  &  \multicolumn{1}{c}{$p <$ }  &     \multicolumn{1}{c} {$p <$ }   \\
          		\midrule
1000  & \textbf{3.5e-19}                 & \textbf{1.7e-32} & \textbf{1.7e-39} & \textbf{1.6e-11} & \textbf{1.6e-32} & \textbf{1.7e-39} \\
2000  & \textbf{3.7e-41}                 & \textbf{6.2e-31} & \textbf{4.0e-41} & \textbf{5.6e-38} & \textbf{2.8e-25} & \textbf{4.0e-41} \\
3000  & \textbf{6.7e-54}                 & \textbf{3.3e-54} & \textbf{3.3e-54} & \textbf{2.0e-52} & \textbf{3.3e-54} & \textbf{3.3e-54} \\
4000  & \textbf{4.6e-51}                 & \textbf{6.1e-51} & \textbf{6.1e-51} & \textbf{4.6e-51} & \textbf{6.3e-51} & \textbf{6.1e-51} \\
5000  & \textbf{7.0e-46}                 & \textbf{1.1e-47} & \textbf{1.1e-47} & \textbf{9.5e-48} & \textbf{1.1e-47} & \textbf{1.1e-47} \\
6000  & \textbf{9.2e-39}                 & \textbf{4.9e-46} & \textbf{4.9e-46} & \textbf{2.5e-44} & \textbf{4.9e-46} & \textbf{4.9e-46} \\
7000  & \textbf{8.5e-22}                 & \textbf{3.2e-45} & \textbf{3.2e-45} & \textbf{4.0e-18} & \textbf{3.2e-45} & \textbf{3.2e-45} \\
8000  & \textbf{2.6e-31}                 & \textbf{6.5e-44} & \textbf{6.5e-44} & \textbf{3.2e-36} & \textbf{6.5e-44} & \textbf{6.5e-44} \\
9000  & \textbf{1.6e-30}                 & \textbf{2.2e-45} & \textbf{2.2e-45} & \textbf{4.8e-44} & \textbf{2.2e-45} & \textbf{2.2e-45} \\
10000 & \textbf{1.3e-25}                 & \textbf{6.1e-51} & \textbf{6.1e-51} & \textbf{1.5e-50} & \textbf{6.1e-51} & \textbf{6.1e-51} \\
		\bottomrule

\end{tabular}
\end{table}

\begin{table}[h!]
	\centering
	\caption{Wilcoxon test ($p$-values) per every length of generated trajectories of the distance to trajectory end point and distance to trajectory middle point feature. Significant differences ($p < 0.05$) are highlighted. $F_A$ = Feature-Based A*, $A_A$ = Attraction-Based A*, $R$ = Real Trajectories, d\_t\_e\_p: distance to trajectory end point, d\_t\: real length trajectories (time-steps)}
	\label{table:statistical_test_d}
\begin{tabular}{ll@{\hskip 0.1in}l@{\hskip 0.1in}l@{\hskip 0.1in}l@{\hskip 0.1in}l@{\hskip 0.1in}l}
		\toprule

      & \multicolumn{3}{c}{d\_t\_e\_p}                             & \multicolumn{3}{c}{d\_t}   \\
      & $F_A\: vs\: A_A$    & $F_A\: vs\: A_A$      & $A_A\: vs\: R$      & $F_A\: vs\: A_A$  & $F_A\: vs\: R$ &      $A_A\: vs\: R$     \\
      		\midrule
    & \multicolumn{1}{c}{$p <$ }   &  \multicolumn{1}{c}{$p <$ }    &  \multicolumn{1}{c}{$p <$ }    & \multicolumn{1}{c}{$p <$ }  &  \multicolumn{1}{c}{$p <$ }  &     \multicolumn{1}{c} {$p <$ }   \\
          		\midrule
1000  & \textbf{1.1e-23}                 & \textbf{2.3e-34} & \textbf{1.7e-39} & \textbf{0.0001}  & \textbf{5.5e-33} & \textbf{1.7e-39} \\
2000  & \textbf{3.7e-41}                 & \textbf{2.5e-35} & \textbf{4.0e-41} & 0.0894           & \textbf{0.0193}  & \textbf{4.0e-41} \\
3000  & \textbf{3.7e-52}                 & \textbf{3.3e-54} & \textbf{3.3e-54} & \textbf{2.8e-06} & \textbf{3.3e-54} & \textbf{3.3e-54} \\
4000  & \textbf{4.7e-49}                 & \textbf{6.1e-51} & \textbf{6.1e-51} & \textbf{4.3e-11} & \textbf{6.3e-51} & \textbf{6.1e-51} \\
5000  & \textbf{9.2e-48}                 & \textbf{1.1e-47} & \textbf{1.1e-47} & \textbf{8.9e-07} & \textbf{1.1e-47} & \textbf{1.1e-47} \\
6000  & \textbf{1.2e-37}                 & \textbf{4.9e-46} & \textbf{4.9e-46} & \textbf{4.1e-06} & \textbf{4.9e-46} & \textbf{4.9e-46} \\
7000  & \textbf{6.5e-19}                 & \textbf{3.2e-45} & \textbf{3.2e-45} & \textbf{4.1e-16} & \textbf{3.2e-45} & \textbf{3.2e-45} \\
8000  & \textbf{8.1e-24}                 & \textbf{6.5e-44} & \textbf{6.5e-44} & 0.2195           & \textbf{6.5e-44} & \textbf{6.5e-44} \\
9000  & \textbf{3.3e-17}                 & \textbf{2.2e-45} & \textbf{2.2e-45} & \textbf{3.3e-14} & \textbf{2.2e-45} & \textbf{2.2e-45} \\
10000 & \textbf{9.9e-20}                 & \textbf{6.1e-51} & \textbf{6.1e-51} & \textbf{2.7e-16} & \textbf{6.1e-51} & \textbf{6.1e-51} \\
		\bottomrule

\end{tabular}
\end{table}

\subsection{Preference between total distance or attractiveness of the environment}
\begin{figure}[!h]
        \centering
        \includegraphics[scale=0.22]{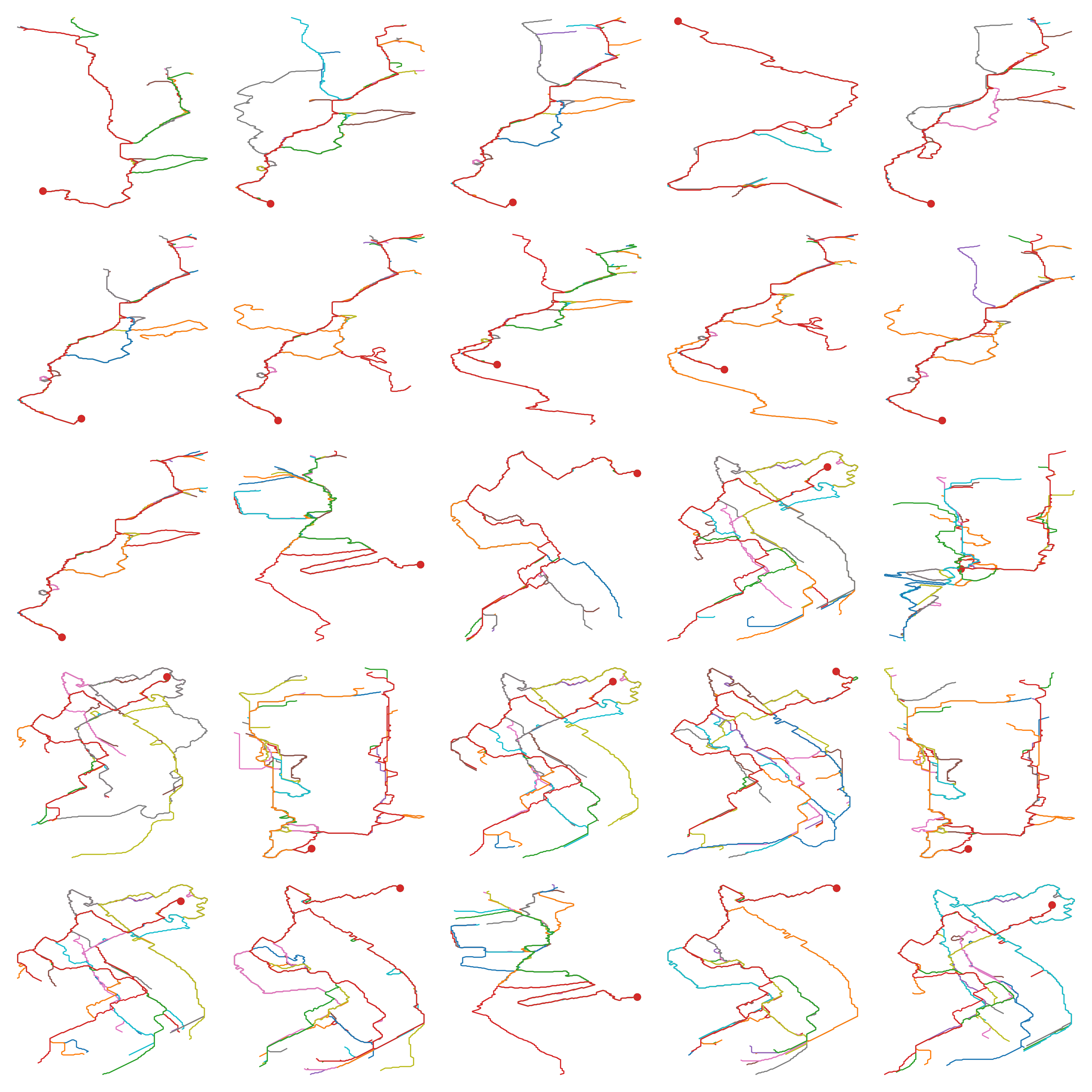}
        \caption{Effect of the alpha parameter on different trajectories generated by the Attraction-Based A*. The starting point is represented by a big circle.}
        \label{fig:results:normal_astar_alpha}
    \end{figure}

\begin{figure}[!h]
        \centering
        \includegraphics[scale=0.22]{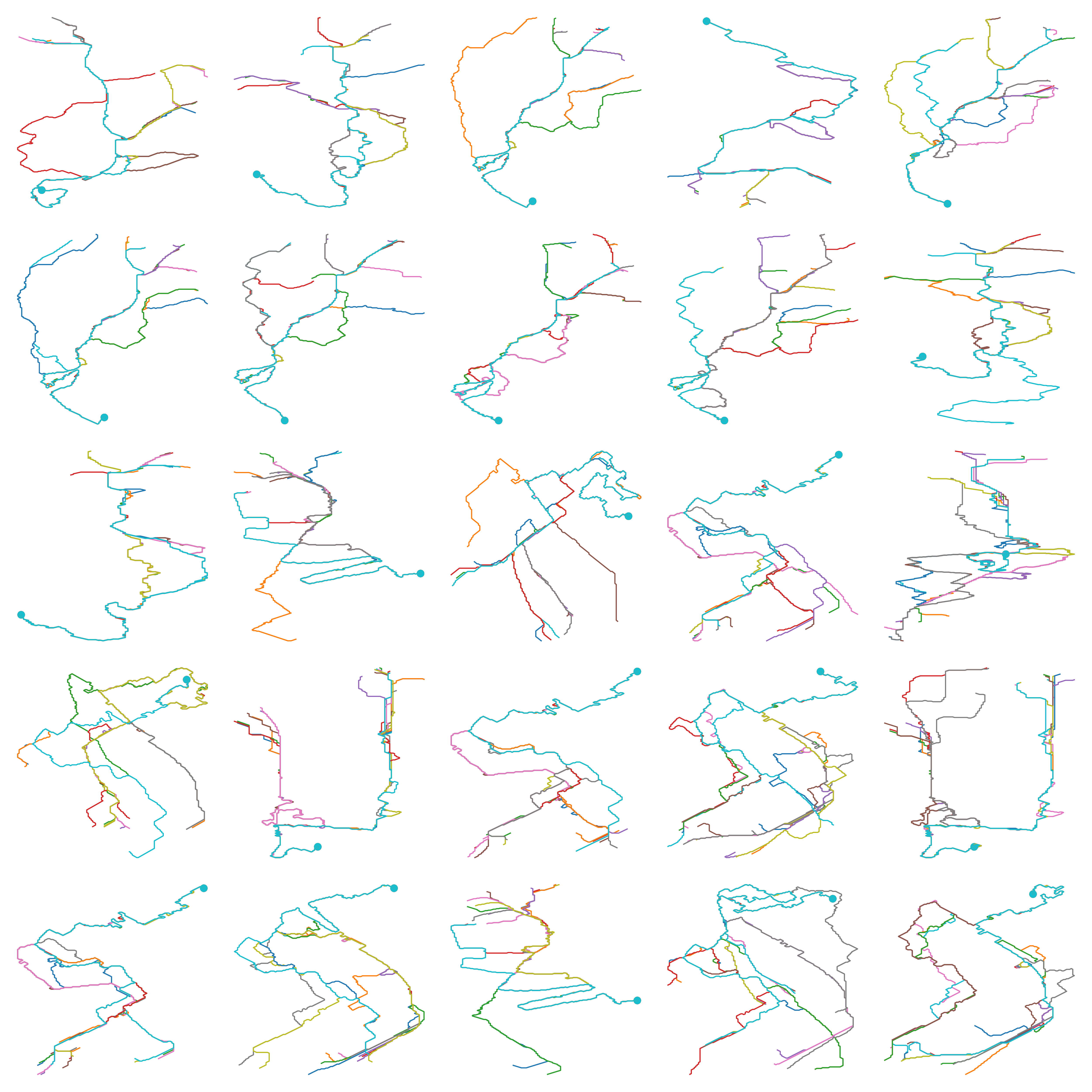}
        \caption{Effect of the alpha parameter on different trajectories generated by the Feature-Based A*. The starting point is represented by a big circle.}
        \label{fig:results:feature_astar_alpha}
    \end{figure}

\begin{figure}
        \centering
        \includegraphics[scale=0.7]{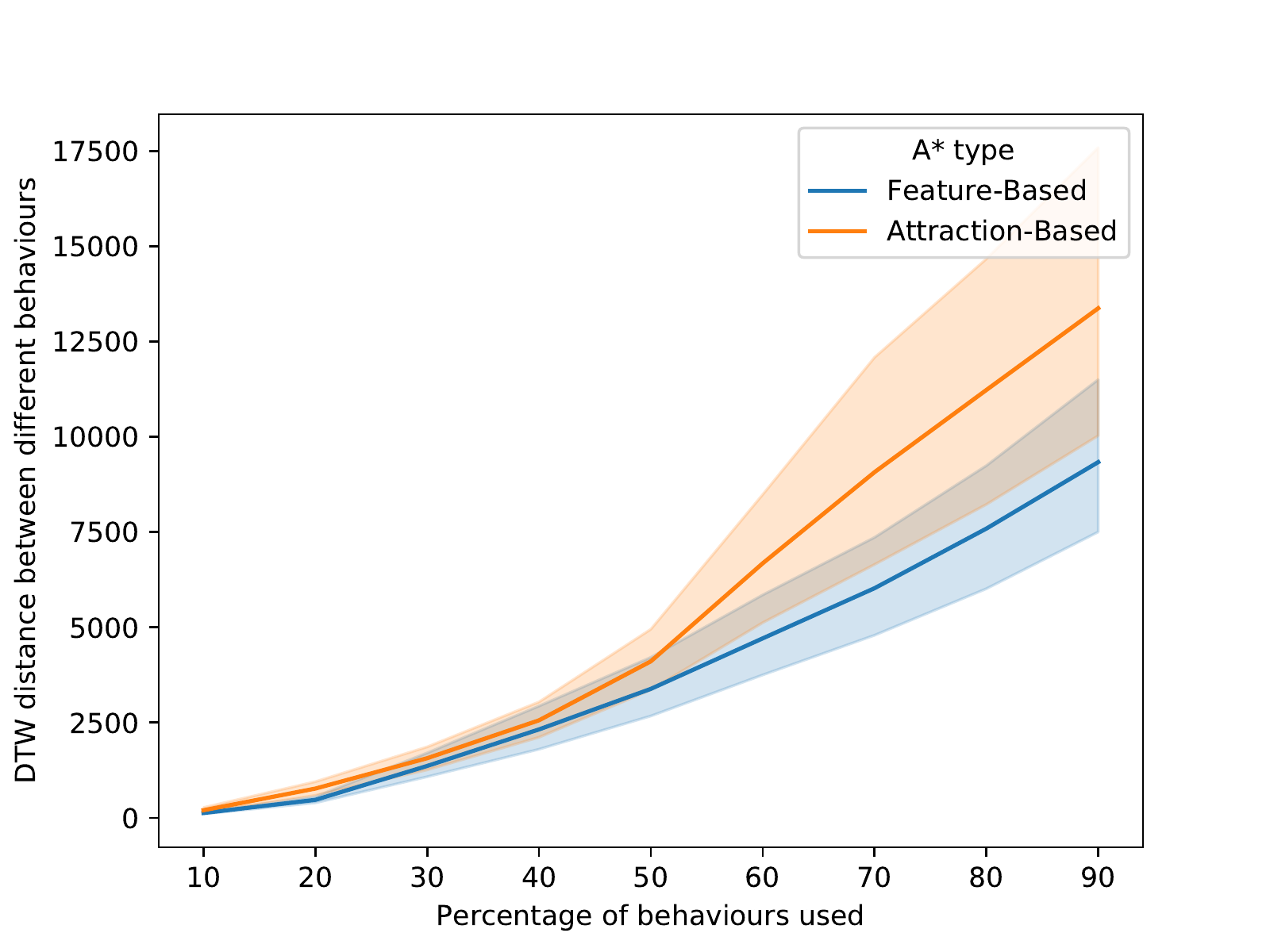}
        \caption{DTW  distance plotted  against  the  percentage  of  distances  considered for the two A* algorithms considered.}
        \label{fig::results:variation_alpha}
    \end{figure}

    The modification of the parameter $\alpha$ in Equation \ref{eq:astar_delta} is very important for the two A*.
    Its tweaking can change how the final trajectories look like.
    Figure \ref{fig:results:normal_astar_alpha} and Fig. \ref{fig:results:feature_astar_alpha}, respectively for Attraction-Based A*  algorithm and Feature-Based A*  algorithm show different trajectories and how they are modified by different $\alpha$ values.
    The initial point is highlighted by a big dot for readability of the trajectories.
    In view of an easier comparison, the two graphs depict trajectories starting from the same starting points.
    Clearly, the initial point is fundamental to notice the effect of the parameters, in other words, the degree of variation in a trajectory depends on how the environment is constituted in the surrounding of the starting point.
    Trajectories behave differently, with some of them showing high variation only in the last section, meanwhile, others show a high degree of variation all along the trajectory.
    This observation can be explained by the location where the starting point is located.
	If the area is full of POIs, and therefore very attractive, the variation of the parameter $\alpha$ has great influence in the way the A*  algorithm computes the next point.
    Figure \ref{fig::results:variation_alpha} shows the two A* algorithms compared in the degree of variation caused by the parameter $\alpha$.
    The DTW distance is plotted against the percentage of distances considered.
    AS visible, in all the cases the Feature-Based A*  algorithm produces less variation than the Attraction-Based A*  algorithm (Wilcoxon signed-rank test $p<0.000$), meaning the Attraction-Based A*  algorithm is easily influenced by the attraction of the environment, more than the Feature-Based A*.

\begin{figure}[!h]
        \begin{subfigure}[b]{0.49\textwidth}
     	 	\centering
     		\includegraphics[scale=0.45]{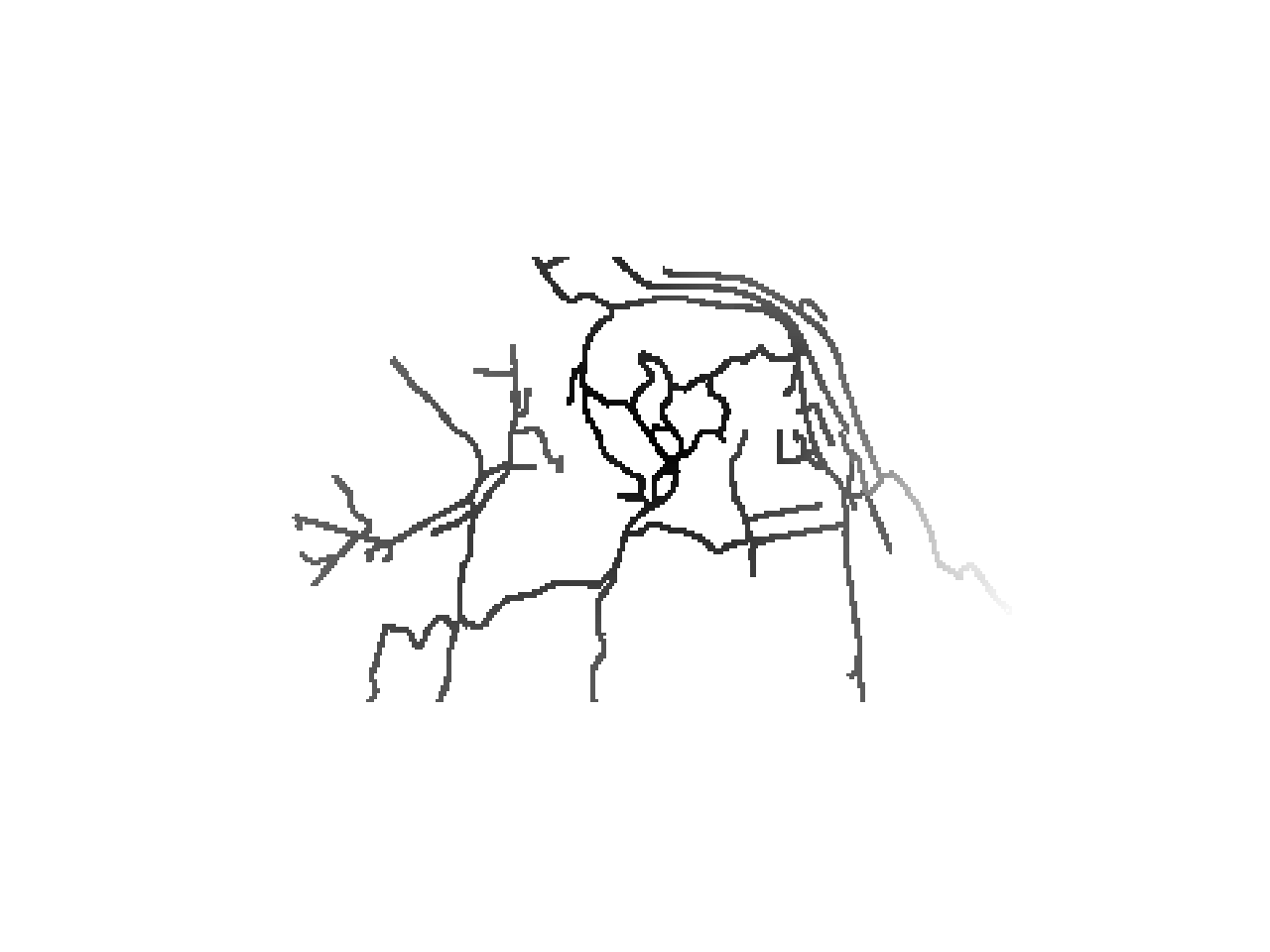}
     		\caption{}
     		\label{img:result:small_matrix_fitness_all_f}
     	\end{subfigure}
     	\begin{subfigure}[b]{0.49\textwidth}
     	 	\centering
     		\includegraphics[scale=0.45]{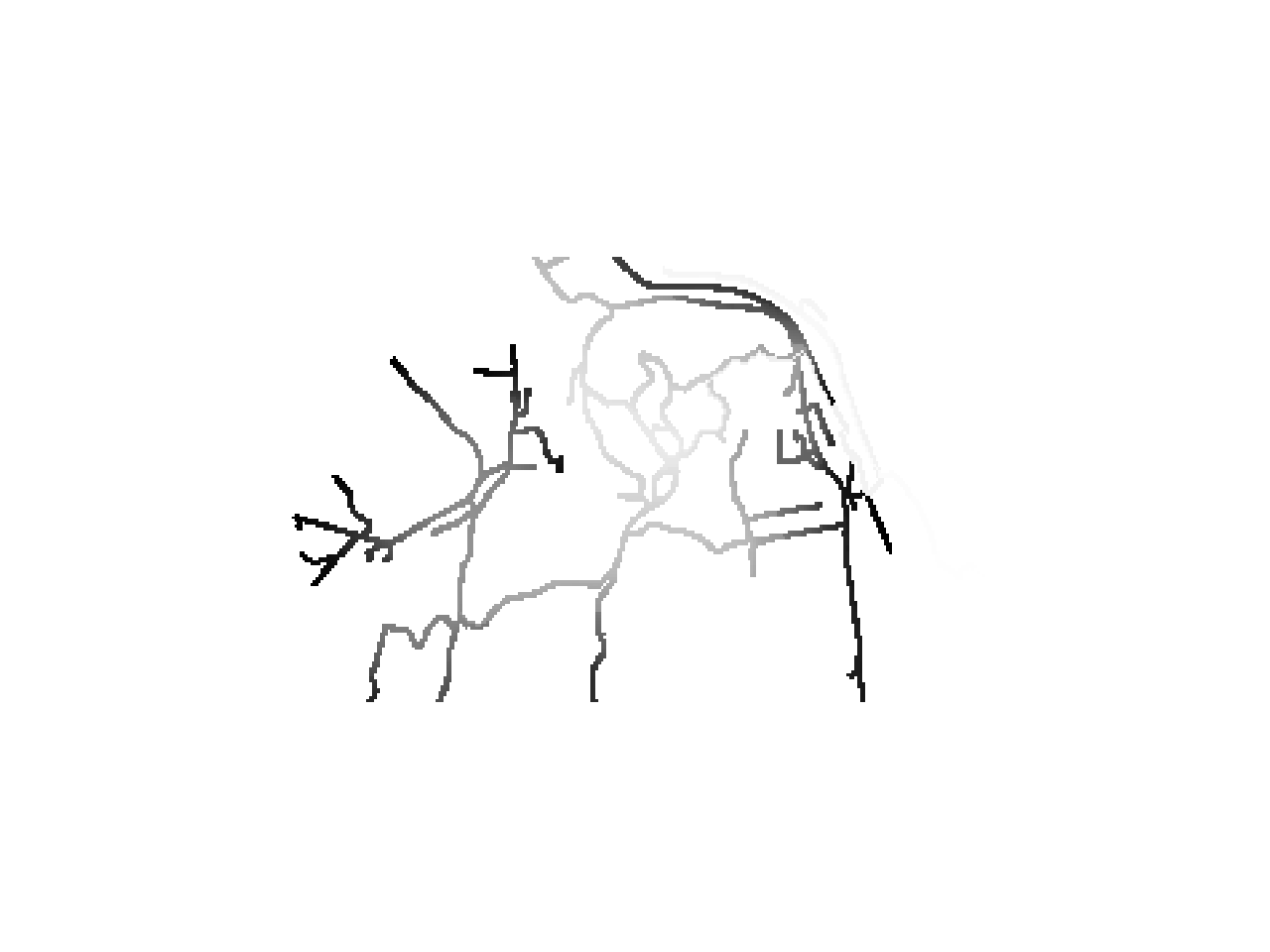}
     		\caption{}
     		\label{img:result:small_matrix_fitness_all_i}
     	\end{subfigure}
         
         \begin{subfigure}[b]{0.49\textwidth}
     	 	\centering
     		\includegraphics[scale=0.45]{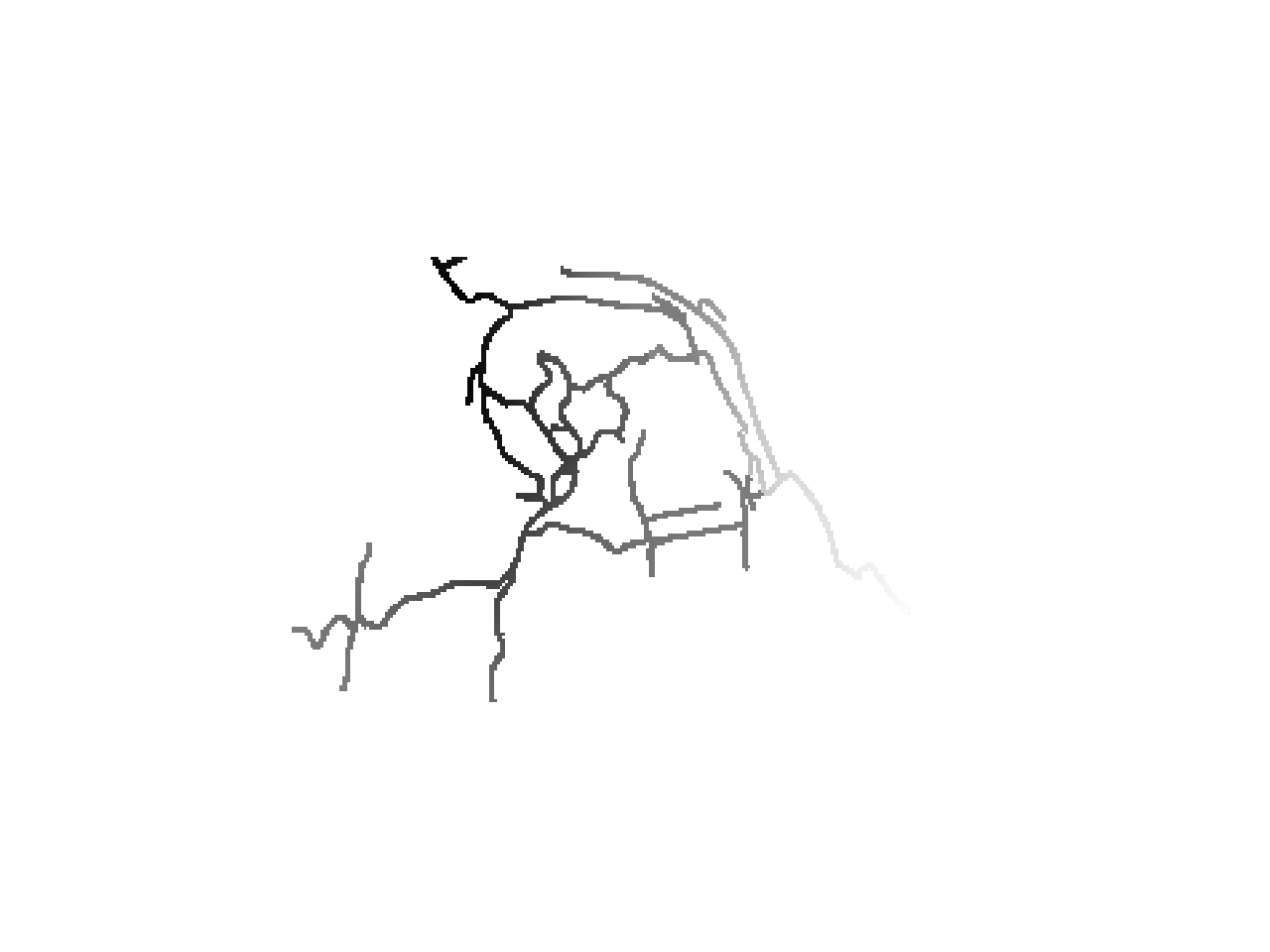}
     		\caption{}
     		\label{img:result:small_matrix_fitness_nothing_f}
     	\end{subfigure}
     	\begin{subfigure}[b]{0.49\textwidth}
     	 	\centering
     		\includegraphics[scale=0.45]{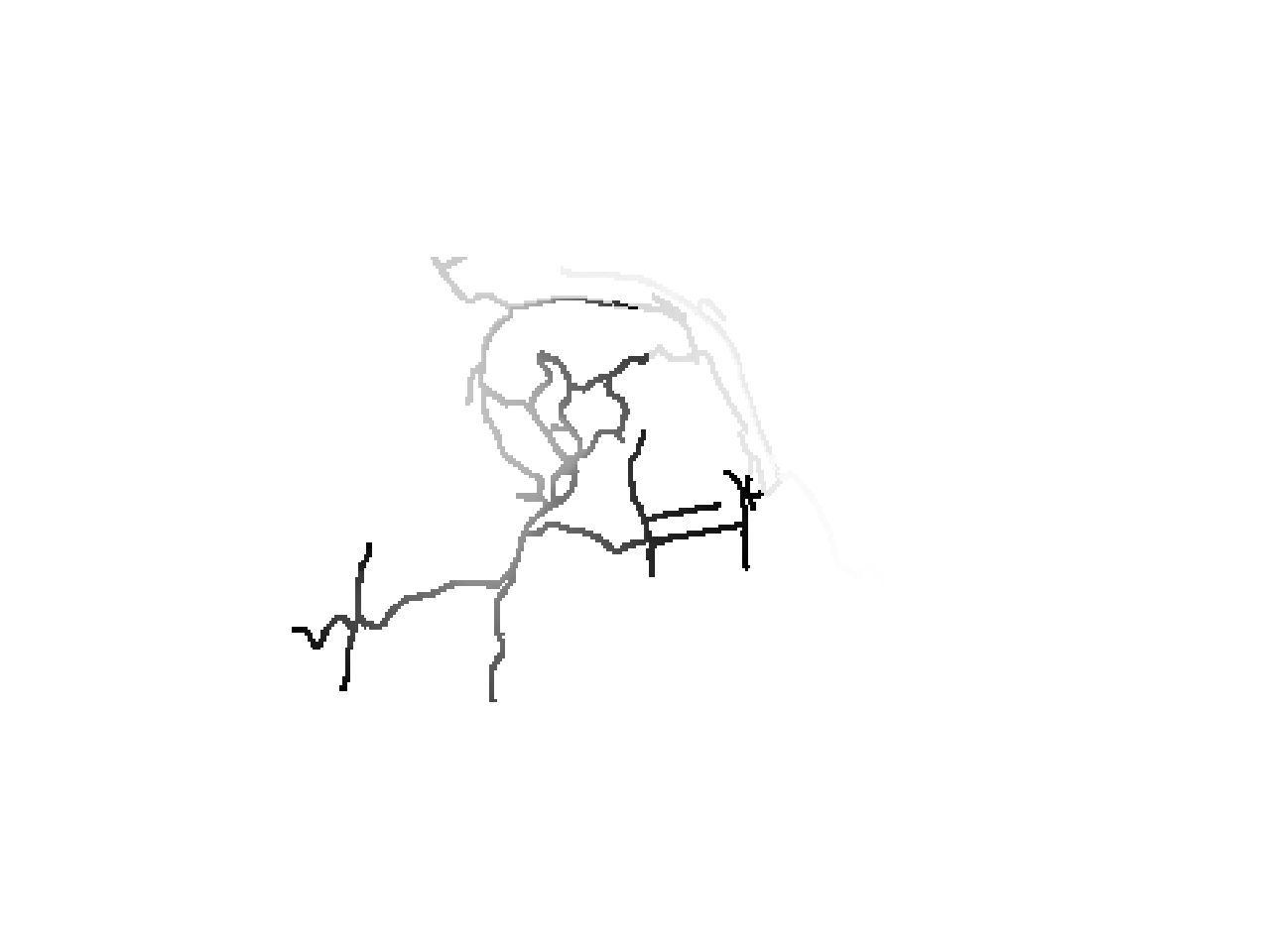}
     		\caption{}
     		\label{img:result:small_matrix_fitness_nothing_i}
     	\end{subfigure}
     	\caption{Raw visualisation of the Attraction-Based A*  algorithm exploring the routing system. Figure \ref{img:result:small_matrix_fitness_all_f} and Fig. \ref{img:result:small_matrix_fitness_nothing_f} show the $f(n)$ value of the A*  algorithm for the max and min $\alpha$ value respectively. The gradient of the colour respect the value of $f(n)$. Figure \ref{img:result:small_matrix_fitness_all_i} and \ref{img:result:small_matrix_fitness_nothing_i} show the progression of the A*  algorithm search for the max and min $\alpha$ value respectively. The gradient of the colour respect the time of the node visited.}
     	 \label{img:result:small_matrix_fitness_raw}
     \end{figure}

    The raw visualisation of how the Attraction-Based A*  algorithm explores the routing system is visible in  Fig. \ref{img:result:small_matrix_fitness_raw}. 
    From the same starting point, the exploration of the A*  algorithm is visualised with the lowest and the highest $\alpha$ values.
    Figure \ref{img:result:small_matrix_fitness_all_f} and Fig. \ref{img:result:small_matrix_fitness_nothing_f} show the $f(n)$ value of the A*, with the gradient of the colour respecting the raw value trend.
    Figure \ref{img:result:small_matrix_fitness_all_i} and \ref{img:result:small_matrix_fitness_nothing_i} show the progression of the A*  algorithm search with the gradient of the colour depicting the time of the node visited.
    Figure \ref{img:result:small_matrix_fitness_all_f} and Fig. \ref{img:result:small_matrix_fitness_all_i} shows the A*  algorithm driven by the distance, with the road going towards the attractive point in the centre but then spreading in all direction looking for the road that fulfils the total distance first. 
    Figure \ref{img:result:small_matrix_fitness_nothing_f} and Fig. \ref{img:result:small_matrix_fitness_nothing_i}, on the other hand, shows the algorithm driven by the attractiveness, exploring more roads around the attractive centre.
    The main difference between the two versions is the amount of road explored by the attraction driven adaptation of the algorithm compared to the distance driven version.
    
\begin{figure}[!h]
        \begin{subfigure}[b]{0.49\textwidth}
     	 	\centering
     		\includegraphics[scale=0.45]{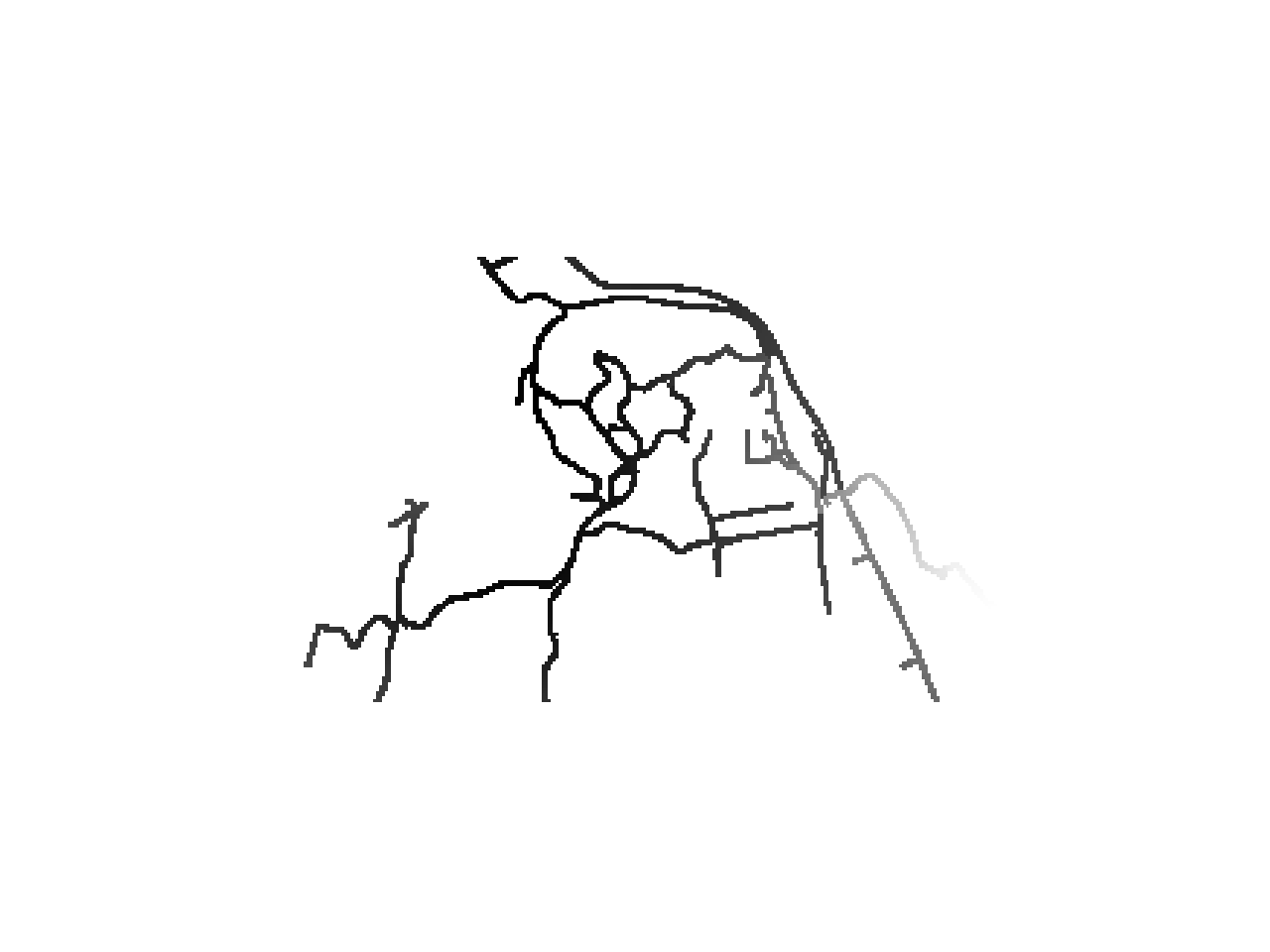}
     		\caption{}
     		\label{img:result:small_matrix_feature_all_f}
     	\end{subfigure}
     	\begin{subfigure}[b]{0.49\textwidth}
     	 	\centering
     		\includegraphics[scale=0.45]{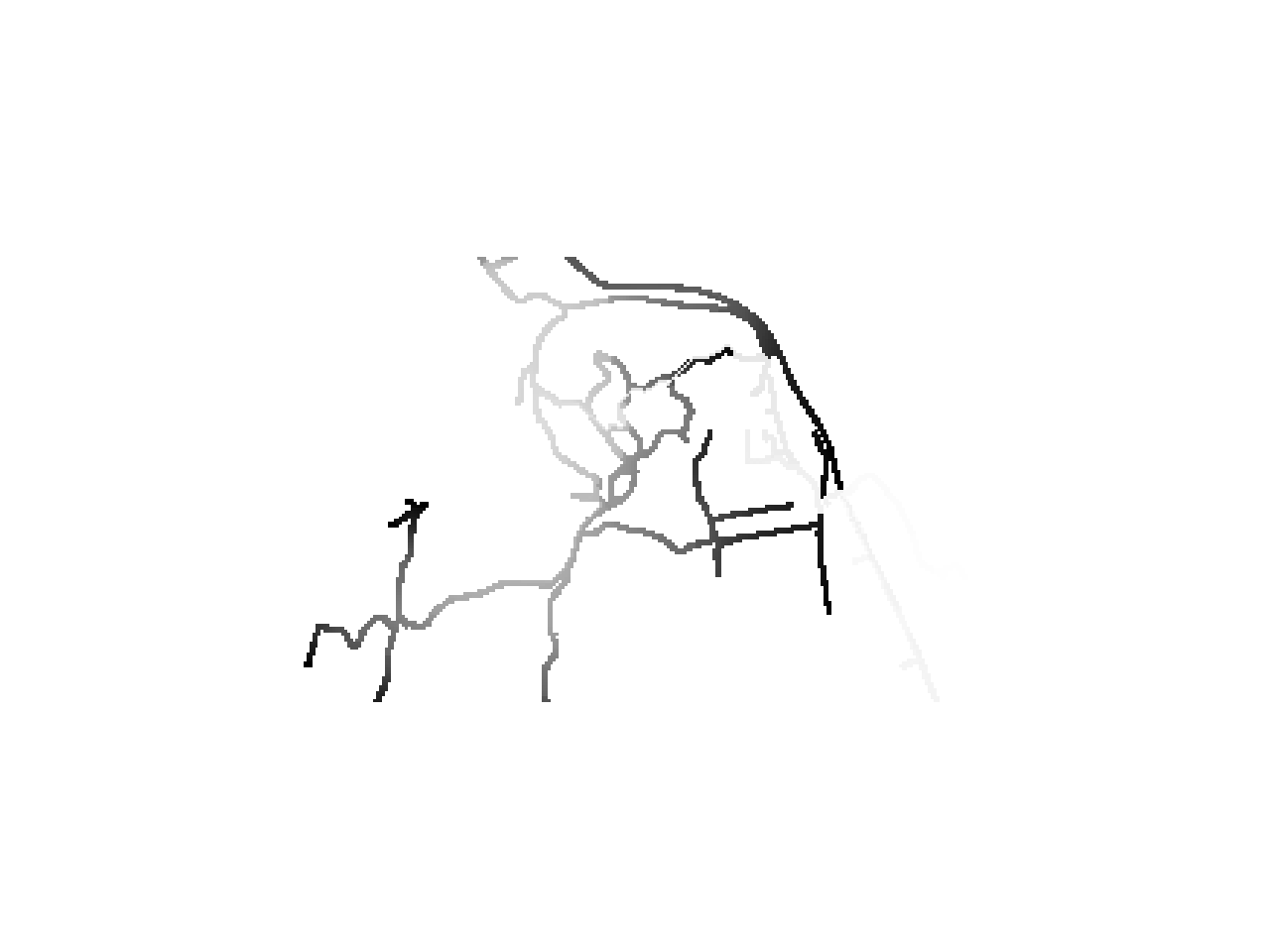}
     		\caption{}
     		\label{img:result:small_matrix_feature_all_i}
     	\end{subfigure}
         
         \begin{subfigure}[b]{0.49\textwidth}
     	 	\centering
     		\includegraphics[scale=0.45]{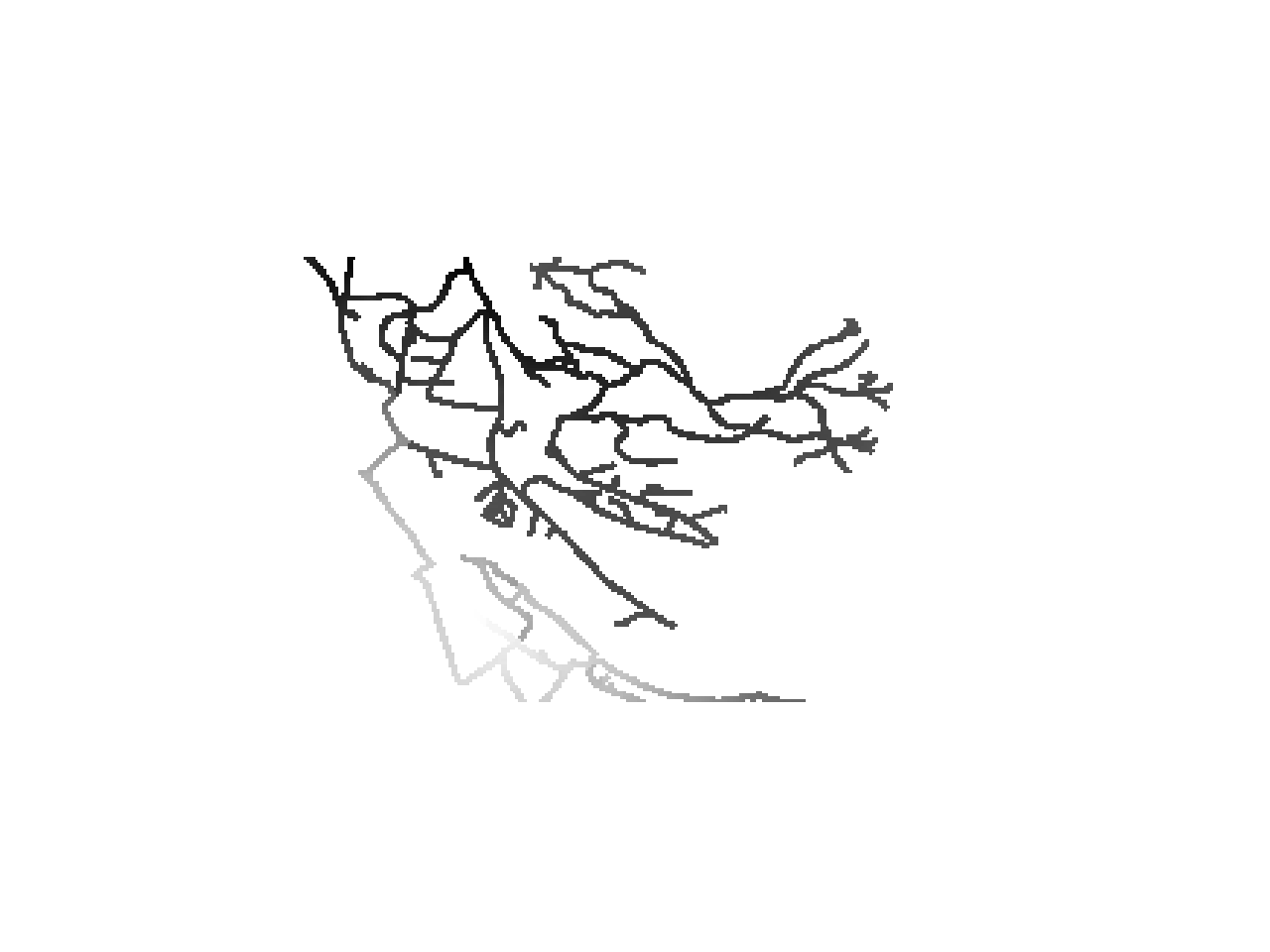}
     		\caption{}
     		\label{img:result:small_matrix_feature_nothing_f}
     	\end{subfigure}
     	\begin{subfigure}[b]{0.49\textwidth}
     	 	\centering
     		\includegraphics[scale=0.45]{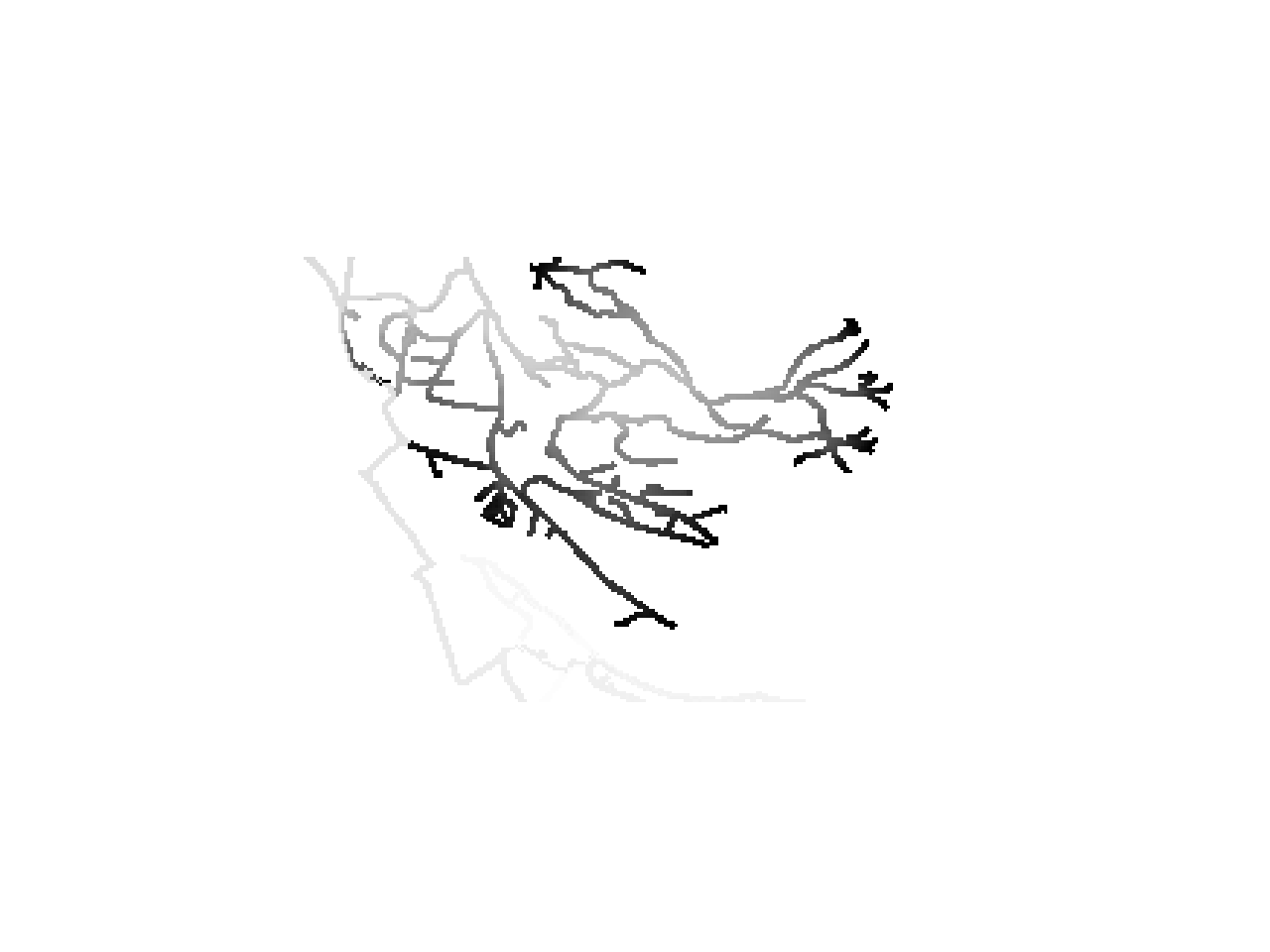}
     		\caption{}
     		\label{img:result:small_matrix_feature_nothing_i}
     	\end{subfigure}
     	\caption{Raw visualisation of the Feature-Based A*  algorithm exploring the routing system. Figure \ref{img:result:small_matrix_feature_all_f} and Fig. \ref{img:result:small_matrix_feature_nothing_f} show the $f(n)$ value of the A*  algorithm for the max and min $\alpha$ value respectively. The gradient of the colour respect the value of $f(n)$. Figure \ref{img:result:small_matrix_feature_all_i} and \ref{img:result:small_matrix_feature_nothing_i} show the progression of the A*  algorithm search for the max and min $\alpha$ value respectively. The gradient of the colour respect the time of the node visited.}
     	 \label{img:result:small_matrix_feature_raw}
     \end{figure}

    Figure \ref{img:result:small_matrix_feature_raw} shows the raw visualisation of how the Feature-Based A*  algorithm explores the routing system.
    From the same starting point, the exploration of the A*  algorithm is visualised with the lowest and the highest $\alpha$ values.
    Figure \ref{img:result:small_matrix_feature_all_f} and Fig. \ref{img:result:small_matrix_feature_nothing_f} show the $f(n)$ value of the A*, with the gradient of the colour respecting its trend.
    Figure \ref{img:result:small_matrix_feature_all_i} and \ref{img:result:small_matrix_feature_nothing_i} show the progression of the A*  algorithm search with the gradient of the colour depicting the time of the node visited.
    As in the example from the Attraction-Based A*, the Feature-Based A*  algorithm when focused in the total distance to generate, it shows exploration from every branch of the reached area, as visible in Fig. \ref{img:result:small_matrix_feature_all_i}. 
    Differently from what visible with the Attraction-Based A*, in this case, fewer roads are explored before returning the final path.
    This observation is not visible in the case with the focus on attraction, where Feature-Based A* algorithm moves to a different area of the routing system compared to the previous instances.
    Here it firstly visits all the roads around the attractive area and subsequentially starts to explore all the branches in order to find the next one, following the structure seen with the version focused on distance.
    
\subsection{Sensitivity to environmental features}
\begin{figure}[!h]
        \centering
        \includegraphics[scale=0.7]{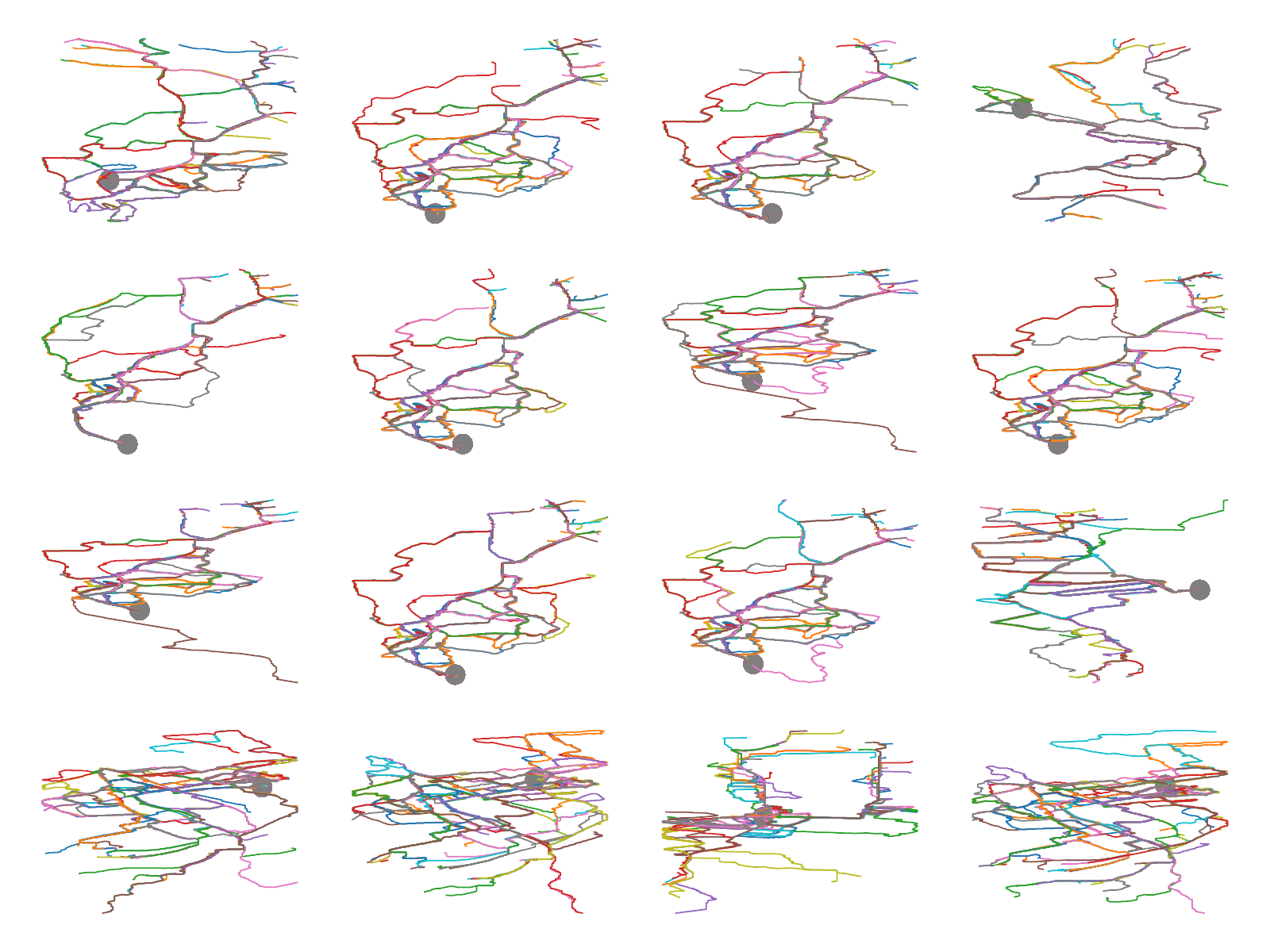}
        \caption{Effect of different environmental attractors on several trajectories generated by the Attraction-Based A*. The starting point is represented by a big circle.}
        \label{fig:results:normal_astar_behaviour}
    \end{figure}

\begin{figure}[!h]
        \centering
        \includegraphics[scale=0.7]{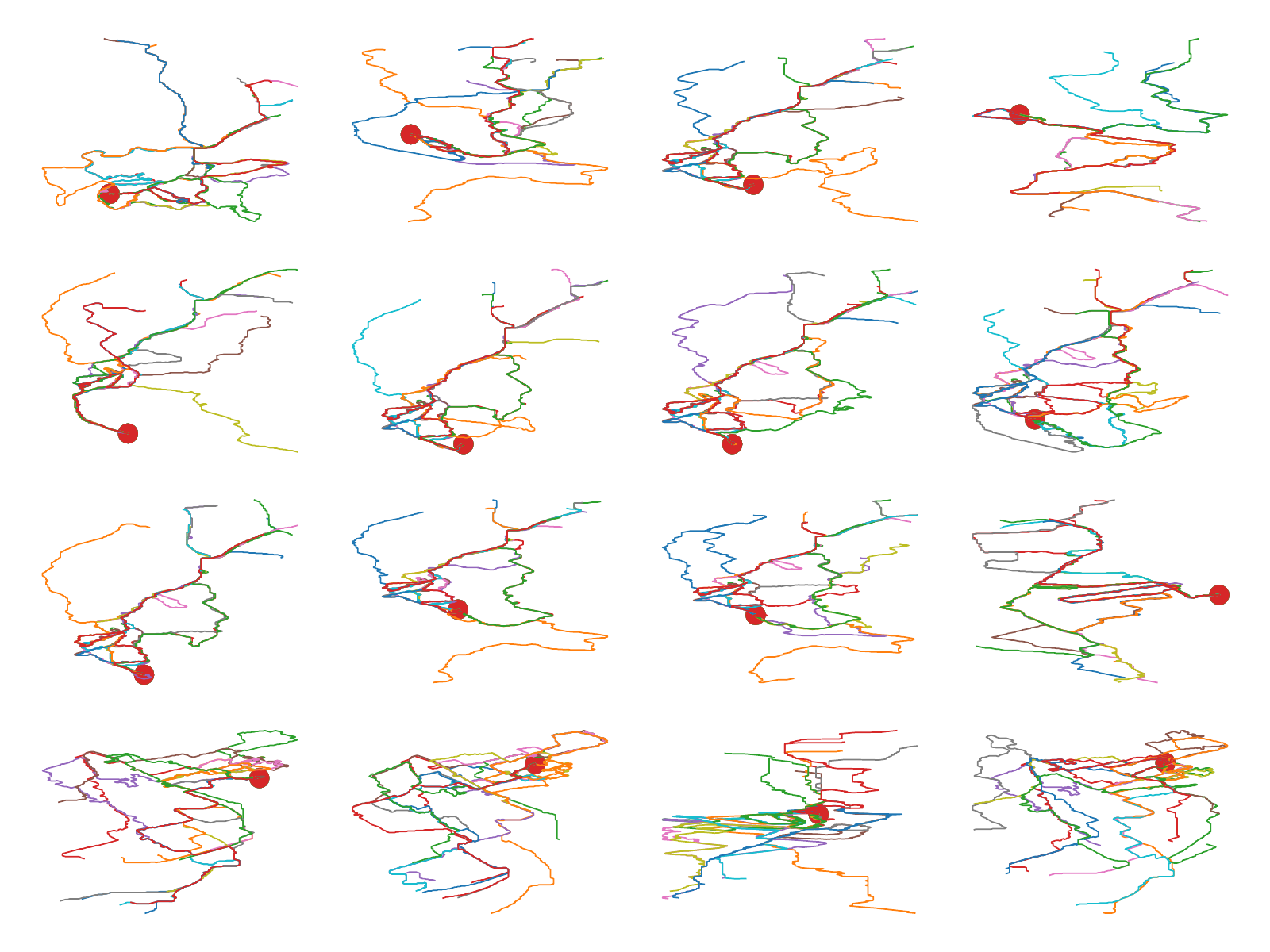}
        \caption{Effect of different environmental attractors on several trajectories generated by the Feature-Based A*. The starting point is represented by a big circle.}
        \label{fig:results:feature_astar_behaviour}
    \end{figure}

\begin{figure}
        \centering
        \includegraphics[scale=0.70]{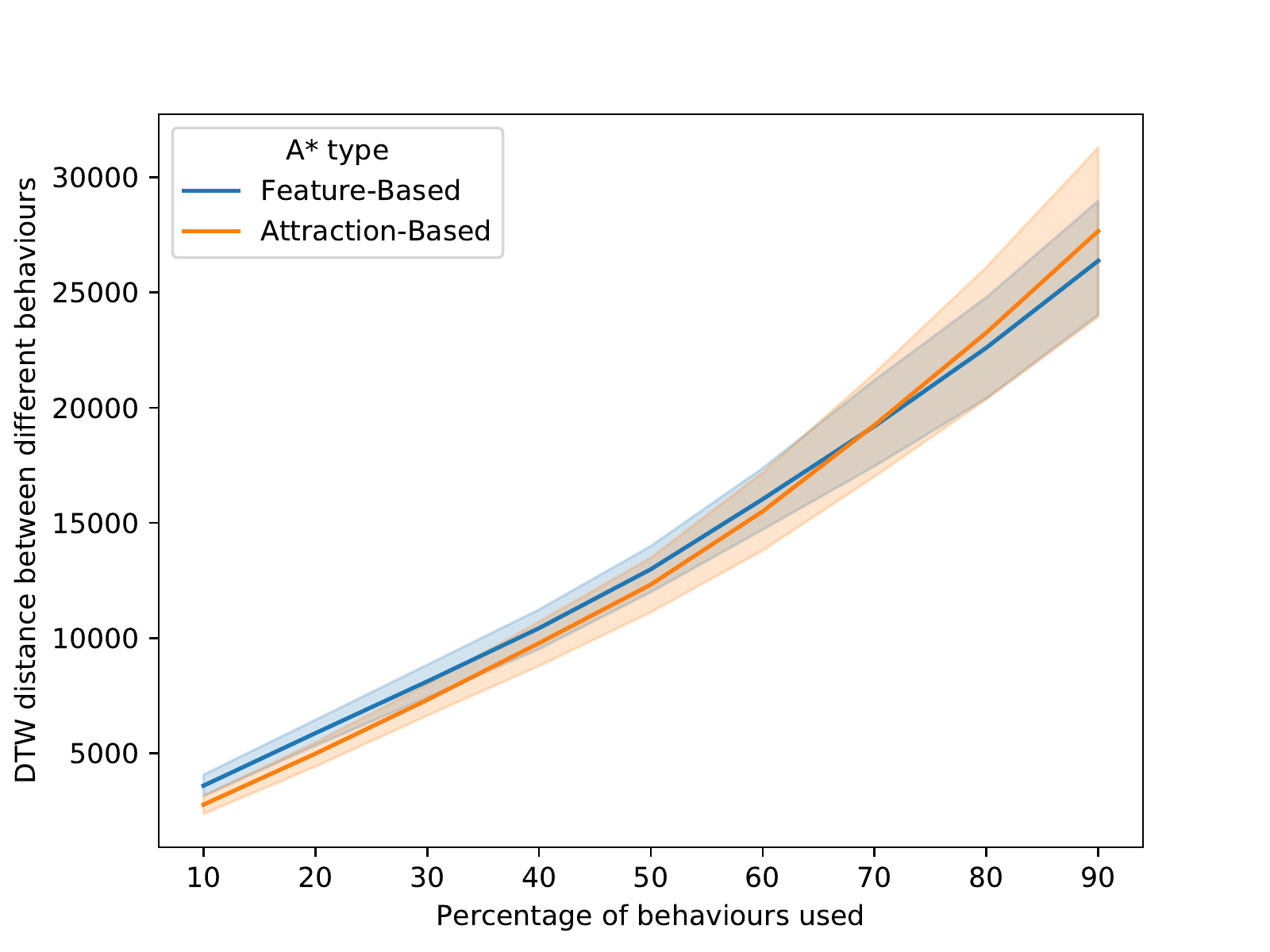}
        \caption{DTW  distance plotted  against  the  percentage  of  distances  considered for the two A* algorithms considered.}
        \label{fig::results:variation_behaviour}
    \end{figure}

    One of the most important aspects of the two A* algorithms is the possibility to tweak the attraction excerpted from all the environmental features to modify the final generated trajectories.
    Figure \ref{fig:results:normal_astar_behaviour} and Fig. \ref{fig:results:feature_astar_behaviour}, respectively for Attraction-Based A*  algorithm and Feature-Based A* algorithm, show different trajectories and how they are modified by tweaking the multipliers.
    The initial point is also highlighted by a big dot for readability of the trajectories.
    The two graphs depict trajectories starting from the same starting points, for sake of comparison.
    The graphs show only a small percentage of the behaviour tested for clarity, and this can lead to some misunderstanding of the effects.
    Some trajectories from Fig. \ref{fig:results:feature_astar_behaviour} seems to generate less variations compared to the same trajectories from Fig. \ref{fig:results:normal_astar_behaviour}.
    Figure \ref{fig::results:variation_behaviour} shows the two A* algorithm compared in the degree of variation caused by different multipliers.
    The DTW distance is plotted against the percentage of distances considered.
    No clear difference between the two algorithms is visible (no statistical difference, Wilcoxon signed-rank test $p=0.09$), which indicates both algorithms are influenced in the same way from the environmental features.
    

\section{Conclusion}
\label{section:conclusion}
We presented an in-depth analysis between two approaches, both able to generate human-like movements based on environmental features.
\cite{Zonta2019} and \cite{Zonta2020} introduced an Attraction and Feature representation of the environment and the real trajectories to incorporate in the algorithm in order to obtain human-like trajectories.  
The human-likeness aspect has been tested by a human expert judging the final generated trajectories, with recognition of almost all the trajectories generated as realistic.
A representation of the world is presented that incorporates environmental features in the map and uses them to influence the process of generating trajectories.

The Efficiency of the two algorithms is compared, with the Attraction-Based A*  algorithm performing incredibly fast compared to the Feature-Based A*, even though under the sampling efficiency it falls short compared to the latest version.
The addition of the feature computations seems to improve the quality of the nodes closed, returning a solution faster than the counterpart, but at a high price, given the amount of computational time needed to compute the components of the feature.

The Feature-Based A*  algorithm is always closer to the real trajectories when the efficacy of the two methods is compared. 
At the cost of computational time, the Feature-Based A*  algorithm is able to be closer to the target values in all the features considered in our comparison.
A weakness of the comparison performed is the maximum length of the trajectories tested, falling short to the real length of the trajectories from the dataset chosen.
A limitation of the Feature-Based A*  algorithm is its slowness that makes impossible to compare the method with the realistic lengths of the trajectories.
Future directions of work would be to study a way to optimise even further the algorithm, in order to flatten its efficiency curve to properly compare the two methods.

The two methods possess several hyper-parameters that can be tweaked to change the final generated trajectories.
One of them is the value $\alpha$ from Equation \ref{eq:astar_delta}, which controls the degree of focusing only on the total distance to generate or on the attractiveness of the environment.
The Feature-Based A*  algorithm produces less diversity in the trajectories generated from the same starting point than the Attraction-Based A*, meaning it gives us less freedom to variate the outcome.
An interesting point for future research would be optimising this value for the optimal resemblance of the generated trajectories to the real ones.

The same outcome is not visible when the other hyper-parameters are tested, specifically the sensitivity to the environmental features.
The two algorithms behave in the same manner, producing a comparable amount of diversity between trajectories starting from the same starting point and conditioned by different environmental feature set.
Interestingly, the diversity caused by different behaviours is higher than the diversity caused by the single value $\alpha$, indicating how powerful is the APF in modifying the environment based on prior information.
Further improvements are possible by increasing the number of tags describing the environment at the expenses of computational power required to precompute the values of the attraction. 
A more fine-grained description of the world,i.e. more descriptors, road conditions such as traffic lights or isolated bike path, might lead to a generation of even more realistic trajectories.

Testing the algorithm on only bike trajectories is a good start, with future research focusing on examining how the system generalises to other modalities, e.g. cars, pedestrians or other means of transportation.

To summarise, we showed how with the injection of information from the real trajectories in the models, we obtain results that are closer to the target, i.e. the real trajectories, but at the expense of computational time. 
Overall, this paper offers an in-depth analysis of existing methods to generate trajectories, to model human
behaviours or even the creation of new datasets with human-like data.

\subsubsection*{Acknowledgements}
The research for this paper was ﬁnancially supported by the Netherlands Organisation for Applied Scientiﬁc Research (TNO)\footnote{\href{https://www.tno.nl/en/}{https://www.tno.nl/en/}}

\bibliographystyle{abbrv}
\bibliography{main.bib}

\end{document}